\documentclass[twoside,journal]{IEEEtran}
\ifCLASSINFOpdf
\else
\fi
\usepackage[usenames, dvipsnames]{xcolor}
\usepackage[colorinlistoftodos]{todonotes}
\usepackage{setspace}
\usepackage{amsfonts}
\usepackage{cite}
\usepackage[hidelinks]{hyperref}
\usepackage{tabularx}
\usepackage{subcaption}
\usepackage{booktabs}
\usepackage{array}
\usepackage{longtable}
\usepackage{pgfgantt}
\usepackage{amsmath}
\usepackage{notoccite}
\usepackage{enumitem}
\usepackage{tikz}
\usetikzlibrary{shapes.geometric, arrows}
\usepackage[export]{adjustbox}
\usepackage{multirow}
\def\layersep{2.5cm}

\newcommand{\fig}[1]  {Fig.\,\ref{#1}}		
\newcommand{\tab}[1]  {Table~\ref{#1}}		
\newcommand{\ssec}[1] {Subsection~\ref{#1}}	

\DeclareMathOperator*{\argmax}{arg\,max}
\begin{document}
%
\title{Differential Evolution and Bayesian Optimisation for Hyper-Parameter Selection in Mixed-Signal Neuromorphic Circuits Applied to UAV Obstacle Avoidance}
%
%
%
%

\author{Llewyn~Salt,
        David~Howard,~\IEEEmembership{Member,~IEEE,}
        Giacomo~Indiveri,~\IEEEmembership{Senior Member,~IEEE,}
        Yulia~Sandamirskaya,~\IEEEmembership{Member,~IEEE,}
\thanks{\IEEEcompsocthanksitem L. Salt is with the School
of Information Technology and Electrical Engineering, University of Queensland, Queensland,
Australia.}%
\thanks{D. Howard is with the Autonomous Systems Lab, CSIRO, Queensland, Australia.}%
\thanks{G. Indiveri and Y. Sandamirskaya are with the Institute of Neuroinformatics, University of Zurich and ETH Zurich, Zurich, Switzerland.
}%
}

%
%

\markboth{Journal of \LaTeX\ Class Files,~Vol.~14, No.~8, August~2015}%
{Shell \MakeLowercase{\textit{et al.}}: Bare Demo of IEEEtran.cls for IEEE Journals}
\maketitle

\begin{abstract}
The Lobula Giant Movement Detector (LGMD) is a an identified neuron of the locust that detects looming objects and triggers its escape responses. Understanding the neural principles and networks that lead to these fast and robust responses can lead to the design of efficient facilitate obstacle avoidance strategies in robotic applications. Here we present a neuromorphic spiking neural network model of the LGMD driven by the output of a neuromorphic Dynamic Vision Sensor (DVS), which has been optimised to produce robust and reliable responses in the face of the constraints and variability of its mixed signal analogue-digital circuits. As this LGMD model has many parameters, we use the Differential Evolution (DE) algorithm to optimise its parameter space. We also investigate the use of Self-Adaptive Differential Evolution (SADE) which has been shown to ameliorate the difficulties of finding appropriate input parameters for DE. We explore the use of two biological mechanisms: synaptic plasticity and membrane adaptivity in the LGMD. We apply DE and SADE to find parameters best suited for an obstacle avoidance system on an unmanned aerial vehicle (UAV), and show how it outperforms state-of-the-art Bayesian optimisation used for comparison.

\end{abstract}


\IEEEpeerreviewmaketitle


%


\section{Introduction}
\IEEEPARstart{S}tate-of-the-art robotic systems are less power efficient and robust than their natural counterparts. Indeed, a bee is capable of robust flight, obstacle avoidance, and cognitive capabilities with a brain that only consumes 10$\mu W$ of power. On the other hand, vehicles in the DARPA Desert and Urban challenges consume around 1$kW$ of power~\cite{Liu2010}. Using nature as inspiration, neuromorphic engineers have attempted to bridge the power-consumption gap through hardware solutions~\cite{Liu2010}. Neuromorphic processors allow for the hardware implementation of spiking neural networks (SNNs)~\cite{chicca2014neuromorphic,indiveri2015neuromorphic}. These mixed-signal analog/digital chips are low power and provide an attractive alternative to current digital hardware used in mobile applications such as robotics.

Another successful neuromorphic solution is the Dynamic Vision Sensor (DVS)~\cite{Lichtsteiner_etal08,Serrano-Gotarredona2013}. The DVS is analogous to a camera, except instead of integrating light in a pixel array for a period of time and then converting it to an image, it detects local changes in luminance at each pixel and transmits these as events pixel by pixel, as they are produced, and with microsecond latency~\cite{delbruck2008frame}. This leads to a reduction in power, bandwidth, and overhead in post processing.

Typically, high-speed agile manoeuvres, such as juggling, pole acrobatics, or flying through thrown hoops use external motion sensors and high powered CPUs to control the UAVs~\cite{Muller2011,brescianini2013quadrocopter,mellinger2011minimum}. A system with sensors and image processing in-situ on the UAV is an essential step for autonomous UAV systems in GPS restricted environments. Due to its high temporal precision, the DVS also does not suffer from blurring as a standard-frame based camera when conducting high-speed manoeuvres on an unmanned aerial vehicle (UAV)~\cite{Mueggler2014}. This makes it ideal as an on-board sensor for high-speed agile manoeuvres.  

A model that has shown promise for collision avoidance in robotics is the locust lobula giant motion detector (LGMD). The locust uses the LGMD to escape from predators by detecting whether a stimulus is looming (increasing in size in the field of view) or not~\cite{Santer2004}. It should be robust to translation, which is why it is an ideal candidate for obstacle avoidance. Previous implementations of this model used frame based cameras and simplified neural models for embedded robotic applications~\cite{Santer2004,yue2010reactive,stafford2007bio}.

Salt et al. \cite{Salt2017} modified the LGMD model to use Adaptive Exponential Integrate and Fire (AEIF) neuron equations which have been shown to be biologically plausible~\cite{brette2005adaptive} and readily implementable in hardware neuromorphic processors\cite{chicca2014neuromorphic}. The LGMD Neural Network (LGMDNN) was also modified to make it compatible with the Reconfigurable On-Line Learning Spiking (ROLLS) neuromorphic processor~\cite{Qiao2015}.  Coupling the LGMDNN with the EIF neural equations yields 11 user-defined parameters after making simplifying assumptions based on the constraints of the neuromorphic processor. Identifying promising parameter sets for robust functional operation of this model is the focus of this work.

Optimising this parameter space is challenging as it contains up to 18 hyper-parameters that have complex inter-dependencies. Due to the computational resources and time requirements involved in evaluation (approximately 1 to 4 minutes per LGMDNN), an exhaustive search is infeasible.  We are therefore motivated to investigate the use of efficient stochastic optimisation algorithms.

Differential Evolution (DE)~\cite{storn1997differential} is particularly suited to our application. DE is a simple and efficient stochastic vector-based real-parameter optimisation algorithm with performance (at least) comparable to other leading optimisation algorithms~\cite{de-vs-pso,de-ga-compare}. DE has only two user-defined rates~\cite{das2011differential,storn1997differential,pedersen2010good}, however their optimal values are problem specific and can drastically affect algorithmic performance~\cite{brest2006self}.  This has prompted research into Self-Adaptation (SA), which allows the rates to vary autonomously in a context-sensitive manner throughout an optimisation run. Self-Adaptive DE (SADE) has been shown to perform at least as well as DE on benchmarking problems~\cite{brest2006self,qin2009differential}.  Importantly, SA has been shown to reduce the number of evaluations required per optimisation in resource-constrained scenarios with protracted evaluation times~\cite{howard2017gecco}, compared to non-adaptive solutions~\cite{howard2015platform}.  Here, we compare DE and SADE to Bayesian Optimisation (BO), which is also well suited to this task.

Spiking networks are particularly amenable to a form of unsupervised learning called Spike-Time Dependent Plasticity (STDP)~\cite{bi-poo}, which allows synaptic weights to change autonomously in response to environmental inputs.  STDP has been shown to provide faster responses compared to non-plastic networks in dynamic environments~\cite{howard2012evolution}, which motivates our investigations into its use in our LGMD networks.

Our hypothesis is that these adaptivity mechanisms are beneficial to the optimisation process.  To test this hypothesis, we evaluate the performance of our algorithms (DE, SADE, and BO, with and without STDP) when optimising looming responses in LGMD networks which are stimulated by (i) simple and (ii) complex DVS recordings on the UAV.

The original contributions of this work are (i) development of an objective function that accurately describes the desired LGMD behaviour, (ii) statistical comparisons of three leading algorithms in optimising LGMD response,  and (iii) the first use of STDP and adaptation in spiking neuromorphic LGMD networks.


\section{Model}
This section will describe the background for the model set-up and the specific equations that were used in the experiment. 
\subsection{LGMD}
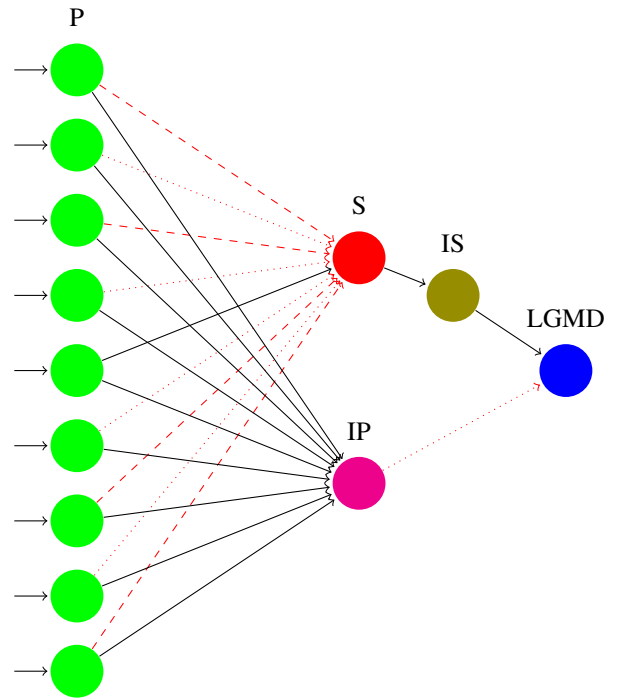
\begin{figure}
\begin{tikzpicture}
	\tikzstyle{excite}=[shorten >=1pt,->,draw=black, node distance=\layersep]
	\tikzstyle{inha}=[shorten >=1pt,->,draw=red,dotted, node distance=\layersep]
	\tikzstyle{inhb}=[shorten >=1pt,->,draw=red,dashed, node distance=\layersep]
    \tikzstyle{every pin edge}=[<-,shorten <=1pt]
    \tikzstyle{neuron}=[circle,fill=black,minimum size=20pt,inner sep=0pt]
    
    \tikzstyle{P neuron}=[neuron, fill=green];
    \tikzstyle{S neuron}=[neuron, fill=red];
    \tikzstyle{IP neuron}=[neuron, fill=magenta];
    \tikzstyle{IS neuron}=[neuron, fill=olive];
    \tikzstyle{LGMD neuron}=[neuron, fill=blue];
    \tikzstyle{annot} = [text width=6em, text centered]

    \foreach \name / \y in {1,...,9}
        \node[P neuron, pin=left:] (I-\name) at (0,-\y) {};

    \node[S neuron] (H) at (1.5*\layersep, -3.5) {};
    \node[IS neuron] (IS) at (2*\layersep,-4) {};
	\node[IP neuron] (IP) at (1.5*\layersep,-6.5) {};
    \node[LGMD neuron] (O) at (2.6*\layersep,-5) {};

\path (I-5) edge[excite] (H) {};
\foreach \name / \y in {2,4,6,8}
	\path (I-\y) edge[inha] (H) {};
\foreach \name / \y in {1,3,7,9}
	\path (I-\y) edge[inhb] (H) {};
\path (H) edge[excite] (IS);
\foreach \name / \y in {1,...,9}
	\path (I-\y) edge[excite] (IP);
\path (IS) edge[excite] (O);
\path (IP) edge[inha] (O);
    \node[annot,above of=H, node distance=0.7cm] (hl) {S};
    \node[annot,above of=I-1, node distance=0.7cm] (Il) {P};
    \node[annot,above of=O, node distance=0.7cm] {LGMD};
    \node[annot,above of=IS, node distance=0.7cm] {IS};
    \node[annot,above of=IP, node distance=0.7cm] {IP}; 
\end{tikzpicture}
\caption{The neuromorphic LGMD model. Solid lines: excitatory connections; Inhibitory connections;   dashed lines:  slower inhibitions;  dotted lines:   faster inhibitions.}
\label{fig:LGMDMe}
\end{figure}

We implement the model as described by Salt et al.~\cite{Salt2017}. The LGMD model consists of a photoreceptor (P), a summing layer (S), an intermediate photoreceptor (IP), an intermediate summing layer (IS), and an LGMD neuron layer. The intermediate layers can be seen as analagous to sum-pooling layers in deep convolutional neural networks~\cite{liu2015treasure,babenko2015aggregating,fernando2017rank}. These layers are connected by  excitatory (E), inhibitory (I), and feed-forward (F) connections, which are modelled as AEIF neurons. \fig{fig:LGMDMe} shows the topology of the network \cite{Salt2017}.   

The feed-forward neurons (F) are intended to inhibit translational motion. The inhibitory connections (I) from the photoreceptor to the summing layer inhibit non-looming stimuli. The weights of the inhibitory connections are assigned based on their distance from the central excitatory neuron. This connection configuration spans the P layer like a kernel.

The intermediate layers were added to make the model compatible with the CXQuad neuromorphic processor described in~\cite{indiveri2015neuromorphic}. However, Salt et al.~\cite{Salt2017} found that the addition of the intermediate (sum-pooling) layer before the LGMD neuron increased the performance of the network on all but slow circular stimuli. 

\subsubsection{Adaptive Exponential Integrate and Fire Spiking Networks}
We use Adaptive Exponential Integrate and Fire (AEIF) networks; the respective neuron equations follow (\ref{eq:gerstnerAEIF}) and (\ref{eq:current}):

\begin{equation}
\label{eq:gerstnerAEIF}
\frac{dV}{dt} = \frac{-g_L(V-E_L)+g_L\Delta_T\exp(\frac{V-V_T}{\Delta_T})+I}{C},
\end{equation}

\begin{equation}
\label{eq:current}
I=I_e-I_{iA}-I_{iB}-I_{ad},
\end{equation}
where $C$ is the membrane capacitance, $g_L$ is the leak conductance, $E_L$ is the leak reversal potential, $V_T$ is the spike threshold, $\Delta_T$ is the slope factor, $V$ is the membrane potential, $I_{e}$ is an excitatory current, $I_{ad}$ is the adaptation current, and $I_{iA}$ and $I_{iB}$ describe fast/slow inhibitory current~\cite{brette2005adaptive}. When a spike is detected ($V>V_T$) the voltage resets ($V=V_r$), and the post-synaptic neuron receives a current injection from the pre-neuron firing given by:
\begin{align}
\label{eq:inject}
I_{al} &+= q_{al},\\
I_{ad} &+= b,
\end{align}
where the subscript $l$ corresponds to the post-synaptic layer, $q_{al}$ is the current, $b$ is the spike-triggered adaptation, and  the subscript $a$ refers to either excitation or inhibition. To simplify the model for embedded implementation, inhibitory currents were bound as a ratio of the excitatory current:
\begin{equation}
q_{il(A|B)} = inh(A|B)_l\cdot q_{el},
\end{equation}
where the $(A|B)$ notation indicates either A or B type inhibition. The decay of the excitatory or inhibitory currents is described by:

\begin{equation}
\frac{dI_a}{dt}=\frac{-I_a}{\tau_a},\label{eq:currentDecay}
\end{equation}
where $I_a$ is the current and $\tau_a$ is the time constant for the decay. The subscript $a$ refers to either inhibition or excitation. Finally, the decay of the adaptation current is described by:
\begin{equation}
\frac{dI_{ad}}{dt}=\frac{a(V-E_L)-I_{ad}}{\tau_{ad}},
\end{equation}
where $a$ is the sub-threshold adaptation and $\tau_{ad}$ is the time constant for the decay.

Initially, the adaptation current is set to 0, which serves as a comparative baseline when investigating the use of adaptation. 

\subsection{Spike Time Dependent Plasticity}
Spike Time Dependent Plasticity (STDP) is a realisation of Hebbian learning based on the temporal correlations between pre- and post-synaptic spikes. This synaptic plasticity is thought to be fundamental to adaptation, learning, and information storage in the brain~\cite{song2000competitive,sjostrom2010spike}.

\begin{figure}[htbp]
\centering
\includegraphics[width=1\columnwidth]{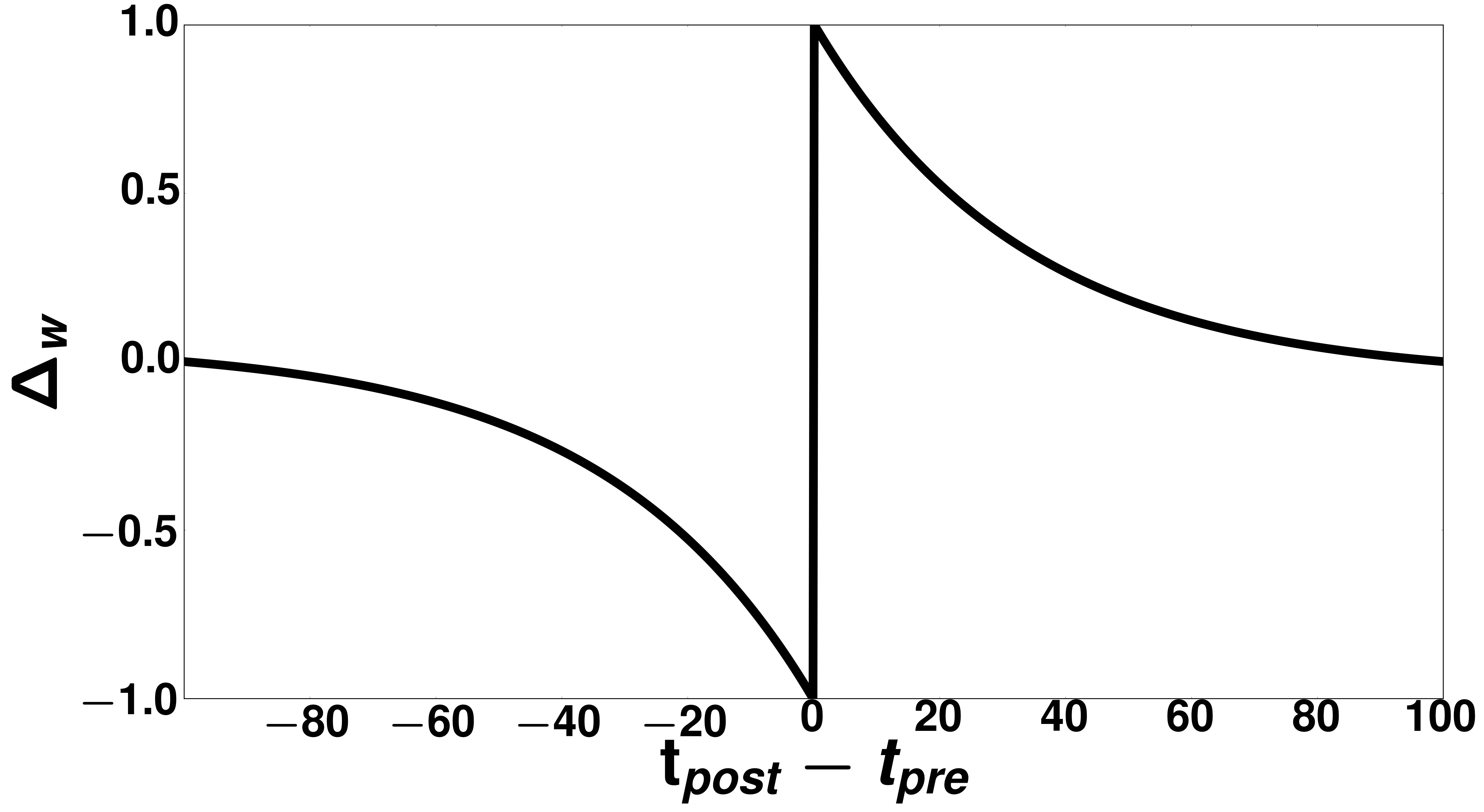}
\caption{The impact of STDP on the synaptic weights. If the pre-synaptic spike arrives before the post synaptic spike, then the strength of the weights is increased. If the post synaptic spike arrives first than the strength of the synapse is weakened.}\label{fig:STDP}
\end{figure}

Considering an arbitrary neuron, receipt of a pre-synaptic spike closely before a post-synaptic spike increases efficacy of the synapse, with the reverse being true if a post-synaptic spike is received in close proximity to a pre-synaptic spike.  Long term potentiating (LTP, synaptic weight increase) of the synapse occurs in the former case, long term depression (LTD, synaptic weight decrease) occurs in the latter case. \fig{fig:STDP} shows the effect of the difference of the post- and pre- synaptic spikes on the synaptic weight. 
STDP modifies the synaptic current injection given in (\ref{eq:inject}) by multiplying it by a weight $w$. If a pre-synaptic spike occurs then:
\begin{align}
I_{al} &+= w q_{al},\\
A_{pre} &+= \Delta_{pre,}\\
w &+= A_{post}.
\end{align}
If a post-synaptic spike occurs then:
\begin{align}
A_{post} &+= \Delta_{post},\\
w &+= A_{pre}.\\
\end{align}
$A_{pre|post}$ are the amount by which the weight $w$ is strengthened or weakened, and $\Delta_{pre|post}$ is a user-defined value for increasing $A_{pre|post}$ each time a spike occurs.
At each spike event:
\begin{align}
\frac{dA_{pre}}{dt} &= -\frac{A_{pre}}{\tau_{pre}},\\
\frac{dA_{post}}{dt} &= -\frac{A_{post}}{\tau_{post}}.
\end{align}
Each time a spike occurs, $A_{pre|post}$ decays according to the function given above. 

\section{Optimisation Techniques}

In this Section, we describe the three optimisation techniques that we compare: DE, SADE, and BO, and how they are applied to optimising the LGMDNN parameter space. Each individual is a parametrisation of the LGMDNN, given by:
$x =$[$\mathbf{\tau_e }$, $\mathbf{\tau_{iA} }$, $\mathbf{\tau_{iB} }$, $\mathbf{q_{eP} }$, $\mathbf{q_{eS} }$, $\mathbf{q_{eIP} }$, $\mathbf{q_{eIS} }$, $\mathbf{q_{eL} }$, $\mathbf{inhA_S}$, $\mathbf{inhB_S}$, $\mathbf{inhA_L}$, [[$\mathbf{a}$, $\mathbf{b}$, $\mathbf{\tau_{w_{adapt}} }$]], (($\mathbf{\tau_{pre} }$, $\mathbf{\tau_{post }}$, $\mathbf{\Delta_{pre}}$, $\mathbf{\Delta_{post}}$))]


\subsection{Differential Evolution}
DE is an efficient and high performing optimiser for real-valued parameters~\cite{das2011differential,storn1997differential}. As it is based on evolutionary computing, it performs well on multi-modal, discontinuous optimisation landscapes.  DE performs a parallel direct search over a population of size $NP$, where each population member $x$ is a $D$-dimensional vector. for each generation $G$:

\begin{equation}
x_{i,G}=1,2,\ldots,NP.
\end{equation}

We use the canonical DE/rand/1/bin to describe the algorithmic process.  The initial population is generated from random samples drawn from a uniform probability distribution of the parameter space, bounded to the range of the respective variable. These bounds are shown in Subsubsection~\ref{sssec:hpc}. The fitness of each vector in the population is then calculated by the objective function, as described in Section~\ref{sec:OF}.

In each generation, each parent generates one offspring by way of a `donor' vector, created following Eq.~(\ref{eq:DErand1bin}):

\begin{equation}
\label{eq:DErand1bin}
v_{i,G+1}=x_{r_1,G}+F\cdot (x_{r_2,G}-x_{r_3,G}),
\end{equation}
where $r_1\neq r_2\neq r_3\neq i \in [1,NP]$ index random unique population members, and differential weight $F\in [0,2]$ determines the magnitude of the mutation.  The final offspring is generated by probabilistically merging elements of the parent with elements of the donor vector. The new vector $u_{i,G+1} = (u_{1i,G+1},\ldots , u_{Di,G+1})$ is found by:

\begin{equation}
u_{ji,G+1}=\begin{cases}
v_{ji,G+1},& \text{if }rand(j)\leq CR \text{ or } j = R,\\
x_{ji,G},& \text{otherwise},
\end{cases}
\end{equation}
where $j \in (1,\ldots,D)$, $CR\in [0,1]$ is the crossover rate, $rand(j)\in [0,1]$ is a uniform random number generator, and $R\in (1,\ldots,D)$ is a randomly chosen index to ensure that at least one parameter changes.  The value of index $i$ is then calculated as:

\begin{equation}
x_{i,G+1} = \begin{cases}
u_{i,G+1},& \text{if }f(u_{ji,G+1})>f(x_{i,G}),\\
x_{i,G},&\text{otherwise}.
\end{cases}
\end{equation}

Once all offspring are generated, they are evaluated on the objective function, and selected into the next generation if they score better than their parent.  Otherwise, the parent remains in the population.

\subsection{Self-Adaptive DE}
Storn and Price~\cite{storn1997differential} showed that DE/rand/1/bin outperformed several other stochastic minimisation techniques in benchmarking tests whilst requiring the setting of only two parameters, $CR$ and $F$.  Many different mutation schemes were subsequently suggested for DE, named following the convention $DE/x/y/z$, where $x$ denotes the vector to be mutated (in this case a random vector), $y$ denotes the number of vectors used, and $z$ denotes the crossover method (bin corresponds to binomial).

Brest et al.~\cite{brest2006self} present the first widely-used self-adaptive rate-varying DE, which is expanded by Qin et al., to allow the mutation scheme to be selected (from four predetermined schemes) alongside the rates~\cite{qin2009differential}, based on previously-successful settings.  Different rates/schemes are shown to work better on different problems, or in different stages of a single optimisation run.  The strategy for a given candidate is selected based on a probability distribution determined by the success rate of a given strategy over a learning period $LP$. A strategy is considered successful when it improves the candidate's value.  In the interest of brevity, we refer the interested reader to~\cite{qin2009differential} for a full algorithmic description.  

Rates are adapted as follows. Before $G>LP$, CR is calculated by randomly selecting a number from a normal distribution, $N(0.5,0.3)$, with a mean of 0.5 and a standard deviation of 0.3. Afterwards it is calculated by a random number from $N(CR_{mk},0.1)$ where $CR_{mk}$ is the median value of the successful $CR$ values for each strategy $K$. $F$ is simply selected from a normal distribution $N(0.5,0.3)$, which will cause it fall on the interval $[-0.4,1.4]$ with a probability of 0.997~\cite{qin2009differential}.


\subsection{Bayesian Optimisation}

Bayesian optimisation (BO), e.g.~\cite{brochu2010tutorial}, is a probabilistic optimisation process that typically requires relatively few evaluations~\cite{mockus1994application, jones2001taxonomy, sasena2002flexibility}, although the evaluations themselves are computationally expensive. When parallelised, BO is shown to locate hyper-parameters within set error bounds significantly faster than other state-of-the-art methods on four challenging ML problems\cite{snoek2012practical}, in one case displaying 3\% improved performance over state-of-the-art expert results. As such, BO can be considered an extremely challenging optimiser as a comparator for DE and SADE, and as SNNs have many hyper-parameters, they are ideal candidates for optimisation.  

BO assumes the network hyper-parameters are sampled from a Gaussian process (GP), and updates a prior distribution of the parameterisation based on observations. For LGMDNN, observations are the measure of generalization performance under different settings of the hyper-parameters we wish to optimise.  BO exploits the prior model to decide the next set of hyper-parameters to sample. 

BO comprises three parts: (i) a prior distribution, (ii) an acquisition function, and (iii) a covariance function.

\subsubsection{Prior}
We use a Gaussian Process (GP) prior, as it is particularly suited to optimisation tasks~\cite{mockus1994application}.  A GP is a distribution over functions specified by its mean, $m$, and covariance, $k$, which are updated as hyper-parameter sets are evaluated.  The GP returns $m$ and $k$ in place of the standard function $f$:

\begin{equation}
f(x)\sim GP(m(x),k(x,x')).
\end{equation}

\subsubsection{Covariance Function}
The covariance function determines the distribution of samples drawn from the GP~\cite{brochu2010tutorial,snoek2012practical}.  Following \cite{snoek2012practical}, we select the 5/2 ARD Mat{\' e}rn kernel (\ref{eq:matern}), where $\theta$ is the covariance amplitude.

\begin{equation}
\label{eq:matern}
k_{m52}(x_i,x_j) = aexp(-\sqrt{5r^2(x_i,x_j)}),
\end{equation}
where:
\begin{equation}
a=\theta(1+\sqrt{5r^2(x_i,x_j)}+\frac{5}{3}r^2(x_i,x_j)),
\end{equation}
where:
\begin{equation}
r^2(x_i,x_j)=\frac{x_i-x_j}{\theta^2}.
\end{equation}

\subsubsection{Acquisition Function}
An acquisition function is a function that selects which point in the optimisation space to evaluate next. We evaluate the three acquisition functions, which select the hyper-parameters for the next experiment: Probability of Improvement (PI), Expected Improvement (EI)\cite{mockus1994application}, and Upper Confidence Bound (UCB)\cite{srinivas2009gaussian} --- see ~\cite{brochu2010tutorial} for full implementation details.  Briefly, the {\bf PI} can be calculated, given our current maximum observation of the GP, $x^+$, by:

\begin{align}
PI(x) &= P(f(x)\geq f(x^+)+\zeta )\nonumber \\ &= \Phi(\frac{\mu(x)-f(x^+)-\zeta}{\sigma(x)}).\label{eq:PIZ}
\end{align}
Here, $\zeta\geq 0$ is a user-defined trade-off parameter that balances exploration and exploitation~\cite{lizotte2008practical}.

{\bf EI} maximises improvement with respect to $f(x^+)$:
\begin{equation}
I(x)=\max\{ 0, f(x)-f(x^+)\}.
\end{equation}
The new sample is found by maximising the expectation of $I(x)$:
\begin{equation}
x = \argmax_x\mathbb{E}(I(x)|\{ \mathbf{X},\mathbf{F}\}).
\end{equation}

Following ~\cite{jones1998efficient},  EI is evaluated by:

\begin{align}
EI(x) &= \begin{cases}
a+\sigma(x)\phi(Z), &\text{if } \sigma(x)>0,\\
0, & \text{otherwise};
\end{cases}\label{eq:EI}\\
a&=(\mu(x)-f(x^+)-\zeta)\Phi(Z);\\\nonumber
Z &= \begin{cases}
\frac{\mu-f(x^+)-\zeta}{\sigma(x)},&\text{if } \sigma(x)>0,\\
0, & \text{otherwise},
\end{cases} \nonumber
\end{align}
where $\phi$ and $\Phi$ correspond to the probability and cumulative distribution functions of the normal distribution, respectively. 

{\bf UCB} maximises the upper confidence bound:
\begin{equation}
UCB(x) = \mu(x) + \kappa \sigma(x),
\end{equation}
where $\kappa \geq 0$ balances exploration and exploitation~\cite{snoek2012practical}, and is calculated per evaluation as: 
\begin{equation}
\kappa = \sqrt{\mathit{\nu \tau_t}},
\end{equation} 
where $\nu$ is the user tunable variable and:
\begin{equation}
\tau_t = 2\log(\frac{t^{\frac{d}{2}+2}\pi^2}{3\delta}).
\end{equation} 
$\delta \in \{0,1\}$, $d$ is the number of dimensions in the function and $t$ is the iteration number.

\section{Test Problem}
This section will outline the rationale of the objective function, the experimental set-up, and assumptions. It is important to note that the motivation behind the model simplifications and objective function is for the work to be directly transferable to the neuromorphic processors described in~\cite{Qiao2015} once they are readily available. 
\subsection{Objective Function}
\label{sec:OF}
The function to optimise was formulated as a weighted multi-objective function~\cite{deb2001multi}. 
We direct the interested reader to~\cite{Salt2016} for a detailed formulation of the objective function, $F_{Acc}(\lambda)$, which is calculated by:

\begin{equation}
F_{Acc}(\lambda) = \begin{cases}
2\times F(\lambda), & \text{if } F(\lambda)>0\text{ and } Acc=1,\\
Acc\times F(\lambda), & \text{if } F(\lambda)>0,\\
0, & \text{if } Acc=1\text{ and }F(\lambda)<0,\\
F(\lambda), & \text{otherwise}.
\end{cases}
\end{equation}
Here, $Acc$ is the accuracy of the LGMDNN output and $F(\lambda)$ is the fitness function. The LGMD network is said to have detected a looming stimulus if the output neuron's spike rate exceeds a threshold $SL$. This can be formalised by:
\begin{equation}
Looming = \begin{cases}
\text{True}, &\text{if } SR>SL,\\
\text{False}, &\text{Otherwise},
\end{cases}
\end{equation}
where $SR$ can be calculated by:
\begin{equation}
SR = \sum^{t+\Delta T}_{i=t} S_i,
\end{equation} 
where $\Delta T$ is the time over which the rate is calculated and $S_i$ is whether or not there is a spike at time $i$; a spike is defined to occur if at time $i$ the membrane potential exceeds $VT$.

The looming outputs are  categorised into true positives ($TP$), false positives ($FP$), true negatives ($TN$), and false negatives ($FN$). Output accuracy is then:
\begin{equation}
Acc = \frac{TP+TN}{TP+TN+FP+FN}.
\end{equation}

$F(\lambda)$ can be calculated by:
\begin{equation}
F(\lambda) = \frac{Score(\lambda)+SSEOS(\lambda)}{2}, 
\end{equation}
where {\em Score} is a scoring function based on the timing of spiking outputs and $SSEOS$ is the sum squared error of the output signal. 

The score is calculated by difference of the penalties' and reward functions' sums over the simulation:
\begin{equation}
Score(\lambda) = \sum^N_{i=1}R_i - \sum^N_{i=1}P_i.
\end{equation}

The reward can at a given time can be calculated by:
\begin{equation}
R(t) = \begin{cases}k\exp(\frac{t}{\Delta t})+1, & \text{if looming and spike},\\
0, & \text{otherwise}.
\end{cases}
\end{equation}

The punishment can be calculated by:
\begin{equation}
P(t) =
\begin{cases}
(l-c)\frac{t}{\Delta t}+c,& \text{if not looming and} \\
\ & \text{spike and } t<\frac{\Delta t}{2};\\
(l-c)\frac{1-(t-\frac{\Delta t}{2})}{\frac{\Delta t}{2}}+c, & \text{if not looming} \\
\ & \text{and spike and } t>\frac{\Delta t}{2};\\
0, & \text{otherwise}.
\end{cases}
\end{equation}
In these equations $t$ and $\Delta t$ remain consistent with the other objective functions and $k$, $l$, and $c$ are all adjustable constants to change the level of punishment or reward.

To calculate $SSEOS(\lambda)$, the signal was first processed so that every spike had the same value. This was done so that the ideal voltage and the actual voltage would match in looming regions, as the voltage can vary for a given spike. Ultimately, the only criterion is that the voltage has crossed the spiking threshold. In the non-looming region the ideal signal was taken to be the resting potential, which was negative for the AEIF model equation. The signal error was calculated at every time step as:

\begin{equation}
SSEOS(\lambda) = -\sum^N_{i=1}(V_{actual}^i-V_{ideal}^i)^2.
\end{equation}

$V_{actual}$ could be obtained directly from the state monitor object of the LGMD output neuron in the SNN simulator (Brian2). $N$ in this case is the length of the simulation and $i$ indicated each recorded data point at each time step of the simulation. $V_{ideal}$ was given by:

\begin{equation}
V_{ideal} = \begin{cases}
V_{spk}, & \text{if looming},\\
V_{r},& \text{otherwise},
\end{cases}
\end{equation}
where $V_{spk}$ is the normalised value given to each spike and $V_r$ is the resting potential.

Overall, this gives an objective function that takes into account the expected spiking behaviour, whilst penalising the system for deviating from plausible voltage values and rewarding it for accurately categorising looming and non-looming stimuli. 

\subsection{Experimental Set-up}
The model was set-up using Brian2 spiking neural network simulator~\cite{goodman2014brian}.

\subsubsection{Data Collection}
Data was collected using a DVS in-situ on a quadrotor UAV (QUAV). Two types of data were collected: simple and real world. The simple data was synthesised using PyGame to generate black shapes on a white background that increased in area in the field of view of the DVS. This included: a fast and slow circle, a fast and slow square, and a circle that loomed then translated while increasing in speed (composite). The laptop playing the stimuli was placed in front of the hovering QUAV and the stimuli were recorded. This was done to maintain any noise that might be generated by the propellers of the QUAV.   

To challenge the model, real stimuli were also recorded: a white ball on a black slope was rolled towards the DVS from 3 different directions; a cup was suspended in the air and the QUAV flew towards and away from the QUAV; and a hand was moved towards and away from the DVS on the hovering QUAV. These are increasing in complexity in terms of the shapes that are presented. 

Two looming and two non-looming events (\~25s) from the composite stimulus were used to optimise the model and then the optimised model was evaluated on the other stimuli. The stimuli were chosen to show that the model generated is both shape and speed invariant.

\subsubsection{Hyper-parameter Constraints}
\label{sssec:hpc}
The hyper-parameters were all continuous and could range from zero to infinity. There were many regions of the parameter space that were not computable even when using a cluster with 368GB of RAM. To mitigate some of the computational difficulties the temporal resolution of the simulation was set to 100$\mu s$. Bayesian optimisation using the expected improvement utility function (BO-EI) was used over 20 eight hour runs to find feasible regions of the optimisation space. 

$C$, $g_L$, $E_L$, $V_T$, and $\Delta_T$ were set as constants as they appeared to have little to no co-dependencies and model performance was not impacted by setting these values and appropriately optimising the other parameters~\cite{Salt2016}: $C=124.2pF$, $g_L=60.05nS$, $E_L=-73.12mV$, $V_T=-3.98mV$, and $\Delta_T=6.71mV$.

\tab{tab:Bounds} shows the constraints found for the rest of the hyper-parameters. 
\begin{table}[htbp]
\centering
\caption{The constraints of the optimisation space}. \label{tab:Bounds}
\begin{tabular}{|>{\raggedright\arraybackslash}p{90pt}||>{\centering\arraybackslash}p{50pt}|>{\centering\arraybackslash}p{50pt}||}
\hline
 \textbf{Parameter}& \textbf{Min} & \textbf{Max} \\
\hline\hline
 $\mathbf{\tau_e}$ & 1 & 10\\\hline
 $\mathbf{\tau_{iA}}$ & 1 & 20\\\hline
 $\mathbf{\tau_{iB}}$ & 1 & 25\\\hline
 $\mathbf{q_{eP}}$ & 1014 & 1363\\\hline
 $\mathbf{q_{eS}}$ & 2000 & 5000\\\hline
 $\mathbf{q_{eIP}}$ & 84   & 230\\\hline
 $\mathbf{q_{eIS}}$ & 119 & 270\\\hline
 $\mathbf{q_{eL}}$ & 29	& 472\\\hline
 $\mathbf{inhA_S}$ &0.04&1.22 \\\hline
 $\mathbf{inhB_S}$ &0.24&1.5 \\\hline
 $\mathbf{inhA_L}$ &0.019&1.3 \\\hline
 $\mathbf{a}$ &1&8\\\hline
 $\mathbf{b}$ &40&141 \\\hline
 $\mathbf{\tau_{w_{adapt}}}$ &1&150 \\\hline
 $\mathbf{\tau_{pre}}$ &1&25 \\\hline
 $\mathbf{\tau_{post}}$ &1&25\\\hline
 $\mathbf{\Delta_{pre}}$ &1e-16&0.05\\\hline
 $\mathbf{\Delta_{post}}$ &1e-16&0.05\\\hline
\hline
\end{tabular}
\end{table}
\subsubsection{Comparing Optimisers}
SADE, DE, BO-EI, BO-PI, and BO-UCB were evaluated thirty times on the same input stimulus, so that they could be statistically compared using a Mann-Whitney U test.
The input stimulus included a black circle on a white background performing a short translation to the right, followed by a half loom, a full recession, and then a full loom (The first two non-loom to loom transitions of the composite stimulus). The stimulus was selected because it consisted of a 50:50 looming to not looming ratio. The values of the user defined parameters were selected as:
\begin{itemize}
\item BO-EI and BO-PI: $\zeta=0.01$;
\item BO-UCB: $\kappa=2.576$;
\item DE: $NP=\frac{10dim}{3}$, $F=0.6607$, $CR=0.9426$;
\item SADE: $LP=3$, $NP=\frac{10dim}{3}$, where dim is the number of hyper parameters.
\end{itemize}\par 
The tests were run using the non-adaptive and non-plastic model with the bounds from \tab{tab:Bounds}. They were defined as having converged if they had not improved for $3\times NP$ evaluations. This meant three generations for the DE algorithms and the same number of BO evaluations. The population size was two more than what is recommended by~\cite{pedersen2010good} for the DE algorithm. This size was chosen as it is relatively small and time was an issue. The short convergence meant that the SADE algorithm needed to have a short LP. The processor time was not included as a metric for this as the tests were run on three different computers so the results would not have been comparable.

\subsubsection{Comparing Models}
Once the best optimiser was found (a comparison of optimisers can be found in \ssec{ssec:optcomp}), the best performing optimiser, SADE, was used to optimise the following models:
\begin{description}
\item[\textbf{LGMD:}] Neuromorphic LGMD; 
\item[\textbf{A:}] LGMD with adaptation;
\item[\textbf{P:}] LGMD with plasticity;
\item[\textbf{AP:}] LGMD with adaptation and plasticity. 
\end{description}

The SADE variables were set to: $LP=3$ and $NP=10dim$. The optimisation process was run 10 times and the best optimiser from these ten runs was selected. The model was then tested on each input case for ten looming to non-looming or non-looming to looming transitions. The performance of each model is reported in \ssec{ssec:modcomp}.

Plasticity was found to degrade the performance sometimes so we experimented clamping it from 0\% to 100\% of the original synaptic strength. This allowed it to range from zero to double the original values when at 100\% to no variation at 0\%.

\section{Results and Discussion}
The results are split into two subsections. First, we will compare the optimisers and then we will compare the addition of adaptation, plasticity, and adaptation and plasticity combined to the baseline model.

The models are evaluated on their accuracy (Acc), sensitivity (Sen), Precision (Pre), and Specificity (Spe). Acc is defined in \ssec{sec:OF}. The other metrics can be found in~\cite{alpaydin}. 
\subsection{Optimiser Comparison and Statistical Analysis}
\label{ssec:optcomp}
\tab{tab:OptimiserResults} shows that the SADE algorithm achieved the best fitness, accuracy, precision, and specificity. The BO-PI algorithm converged on its solution in the least number of objective function evaluations and the DE algorithm achieved the best sensitivity but the worst fitness, precision, and specificity.

\begin{table}[htbp]
\centering
\caption{Optimisation algorithm metrics.\label{tab:OptimiserResults}}
\begin{tabular}{|l||c|c|c|c|c|c||}
\hline
\textbf{Meth} & \textbf{Fit} & \textbf{Eva} & \textbf{Acc} & \textbf{Sen} & \textbf{Pre} & \textbf{Spe} \\
\hline\hline
 \textbf{BPI} & -197.1 & \textbf{162.5} &  0.64 &  0.47 &  0.78 &  0.81 \\
 \textbf{DE} & -675.4 &  238.8 &  0.62 & \textbf{0.65} &  0.76 &  0.59 \\
 \textbf{BEI} & -454.0 &  181.3 &  0.63 &  0.57 &  0.80 &  0.69 \\
 \textbf{SADE} & \textbf{-84.9} &  253.2 & \textbf{0.66} &  0.45 & \textbf{0.88} & \textbf{0.87} \\
 \textbf{BUCB} & -533.3 &  180.0 &  0.62 &  0.61 &  0.78 &  0.64 \\
\hline
\end{tabular}
\end{table}

 \begin{table}[htbp]
\centering
\caption{Comparison of the statistical significance of the results. }\label{tab:statistics}
\begin{tabular}{|l|l||c|c|c|c|c|c||}
\hline
 \textbf{} & \textbf{Meth} & \textbf{Fit} &\textbf{ Eva}& \textbf{Acc} & \textbf{Sen} & \textbf{Pre} & \textbf{Spe} \\
\hline
\hline
\multirow{4}{*}{\textbf{BUCB}} &BPI   & + & .& . & + & .& . \\\cline{2-8}
&DE & . & + & . & . & . & . \\\cline{2-8}
&BEI    & . & . & . & . & . & . \\\cline{2-8}
&SADE   & + & + & . & + & + & + \\\cline{2-8}
\hline  
\hline
\multirow{4}{*}{\textbf{DE}} &BPI  & + & + & . & + & . & . \\\cline{2-8}
 &BEI    & . & + & . & . & . & . \\\cline{2-8}
 &SADE & + & . & + & + & + & + \\\cline{2-8}
 &BUCB  & . & + & . & . & . & . \\\cline{2-8}
\hline    
\hline
\multirow{4}{*}{\textbf{BEI}} &BPI & + & . & . & + & . & . \\\cline{2-8}
 &DE   & . & + & . & . & . & . \\\cline{2-8}
 &SADE & + & + & . & + & . & . \\\cline{2-8}
 &BUCB   & . & . & . & . & . & . \\\cline{2-8}
\hline   
\hline
\multirow{4}{*}{\textbf{SADE}}&BPI   & . & + & . & . & + & + \\\cline{2-8}
 &DE    & + & . & + & + & + & + \\\cline{2-8}
 &BEI    & + & + & . & + & . & . \\\cline{2-8}
 &BUCB   & + & + & . & + & + & + \\\cline{2-8}
\hline  
\hline
\multirow{4}{*}{\textbf{BPI}} &DE    & + & + & . & + & . & . \\\cline{2-8}
 &BEI    & . & . & . & + & . & . \\\cline{2-8}
 &SADE  & . & + & . & . & + & + \\\cline{2-8}
 &BUCB   & + & . & . & + & . & . \\\cline{2-8}
\hline
\hline
\end{tabular}
\end{table}

\tab{tab:statistics} shows the statistical significance of the results from \tab{tab:OptimiserResults}. The method in the comparison column is compared to each method in the subsequent column. A + indicates statistically significant values and a . indicates no statistical significance. Statistical significance was defined as $p\leq 0.05$. The Mann-Whitney U test was used to determine statistical significance because it does not require normally distributed samples.

SADE's better fitness is statistically significant compared to all optimisers other than BO-PI. However, to achieve this fitness it also performed the most evaluations when compared to the others. This difference is significant compared to all the optimisers except for DE, which has almost the same number of evaluations. SADE also has significantly worse sensitivity than all but the BO-PI algorithm. Both SADE and BO-PI scored the best fitness values whilst exhibiting the significantly lowest sensitivity values when compared to the other algorithms.

BO-PI was significantly better than DE and BO-UCB for fitness. It also had significantly less evaluations than DE and SADE. Its precision and specificity is significantly less than  the SADE algorithm.

BO-EI has significantly worse fitness and sensitivity when compared to BO-PI and SADE.

DE took significantly more evaluations to converge when compared to all algorithms but SADE. It also had significantly worse fitness, accuracy, precision and specificity than SADE but significantly higher sensitivity. It had significantly worse fitness but significantly better sensitivity than BO-PI.

BO-UCB had significantly worse fitness but better sensitivity than SADE and BO-PI. It also had significantly worse precision and specificity than SADE.

A possible reason that DE underperformed is that the $F$ values provided in \cite{pedersen2010good} are not appropriate for this problem. The population size may have also been too small. Before the SADE algorithm was implemented, doubling the recommended population size made DE find better results than when it had a smaller population. When the population size is too small, whole regions of the parameter space can be missed resulting in poor performance.

SADE removes the need to find control parameters and has been shown to perform as well or better than DE even when the control parameters are well selected~\cite{qin2009differential}. The generalisability that comes with finding the right control parameters on-the-fly is also appealing.

The addition of the various mutation functions to SADE also seems to help it find better results. This is probably due to the desirable properties of each mutation function cancelling out the undesirable properties of other mutation functions.

A surprising result was that of the BO algorithms BO-PI seemed to perform the best. This is contrary to what the authors in~\cite{brochu2010tutorial} found. They suggested that it tended to have the worst performance of the three.

\subsection{SADE Averages}
\begin{figure}[h!tbp]
\begin{subfigure}{1\columnwidth}
\centering
\includegraphics[width=1\columnwidth]{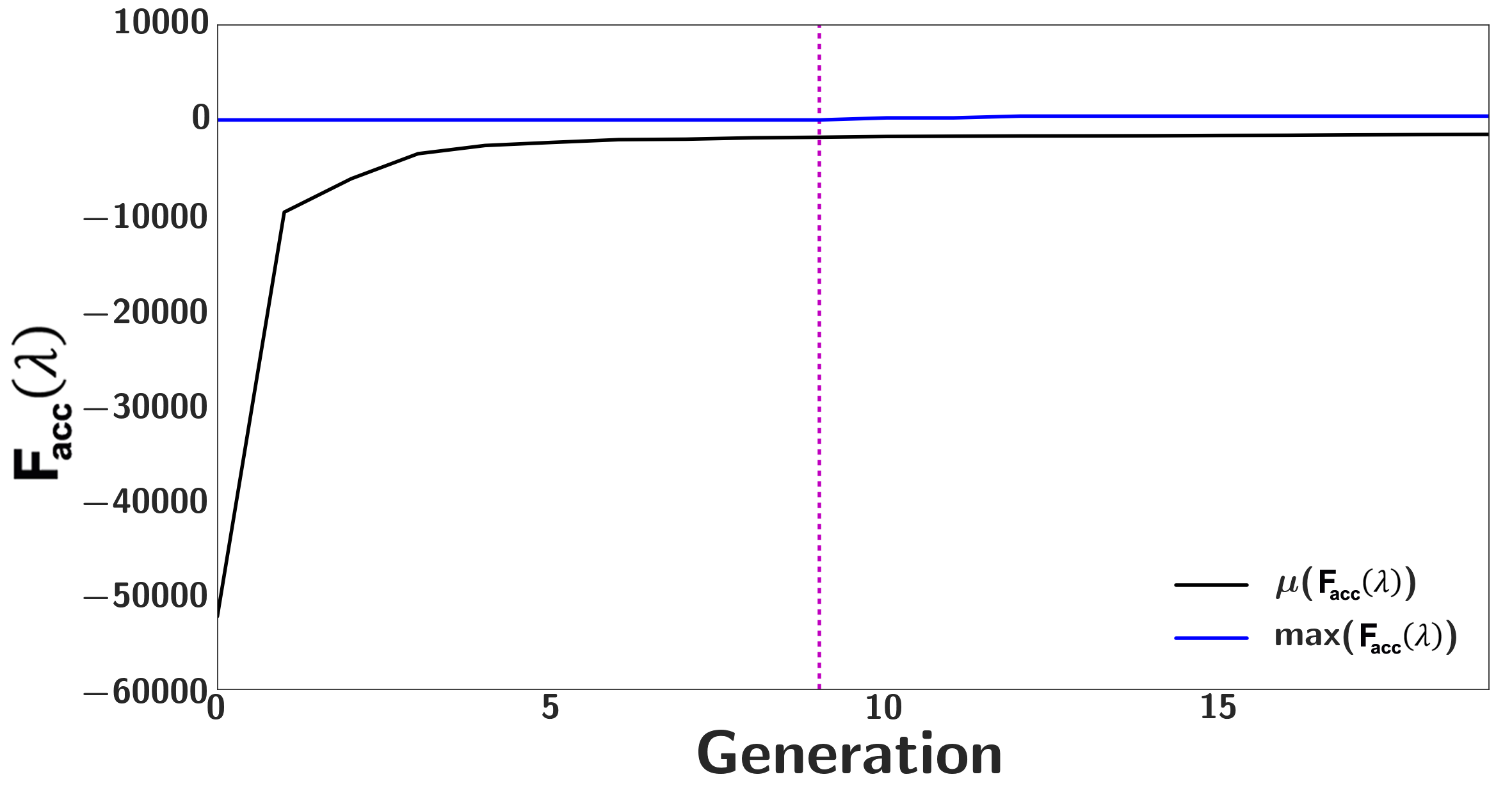}
\caption{}\label{fig:sAVG}
\end{subfigure}
\begin{subfigure}{1\columnwidth}
\centering
\includegraphics[width=1\columnwidth]{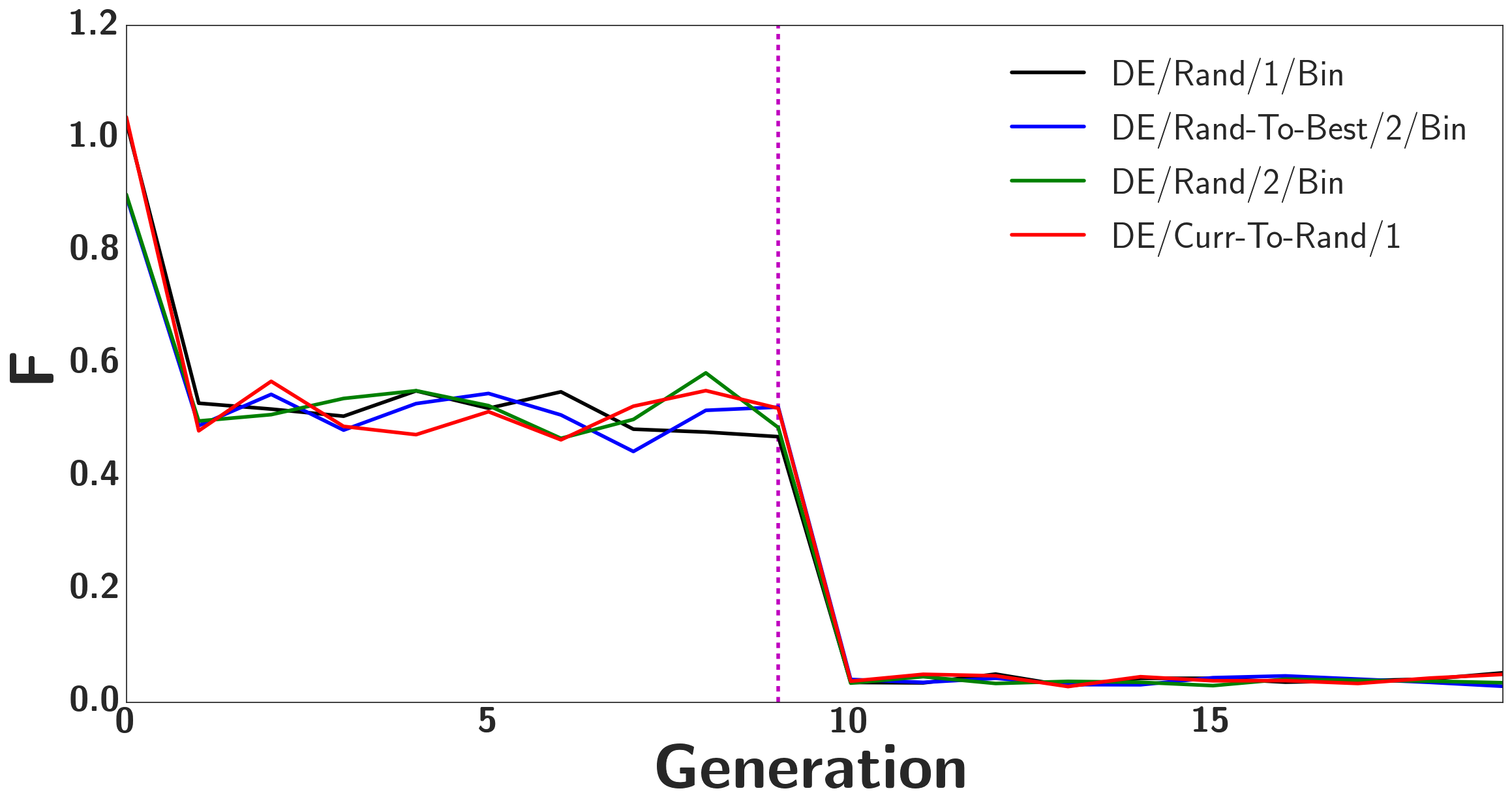}
\caption{}\label{fig:fAVG}
\end{subfigure}
\begin{subfigure}{1\columnwidth}
\centering
\includegraphics[width=1\columnwidth]{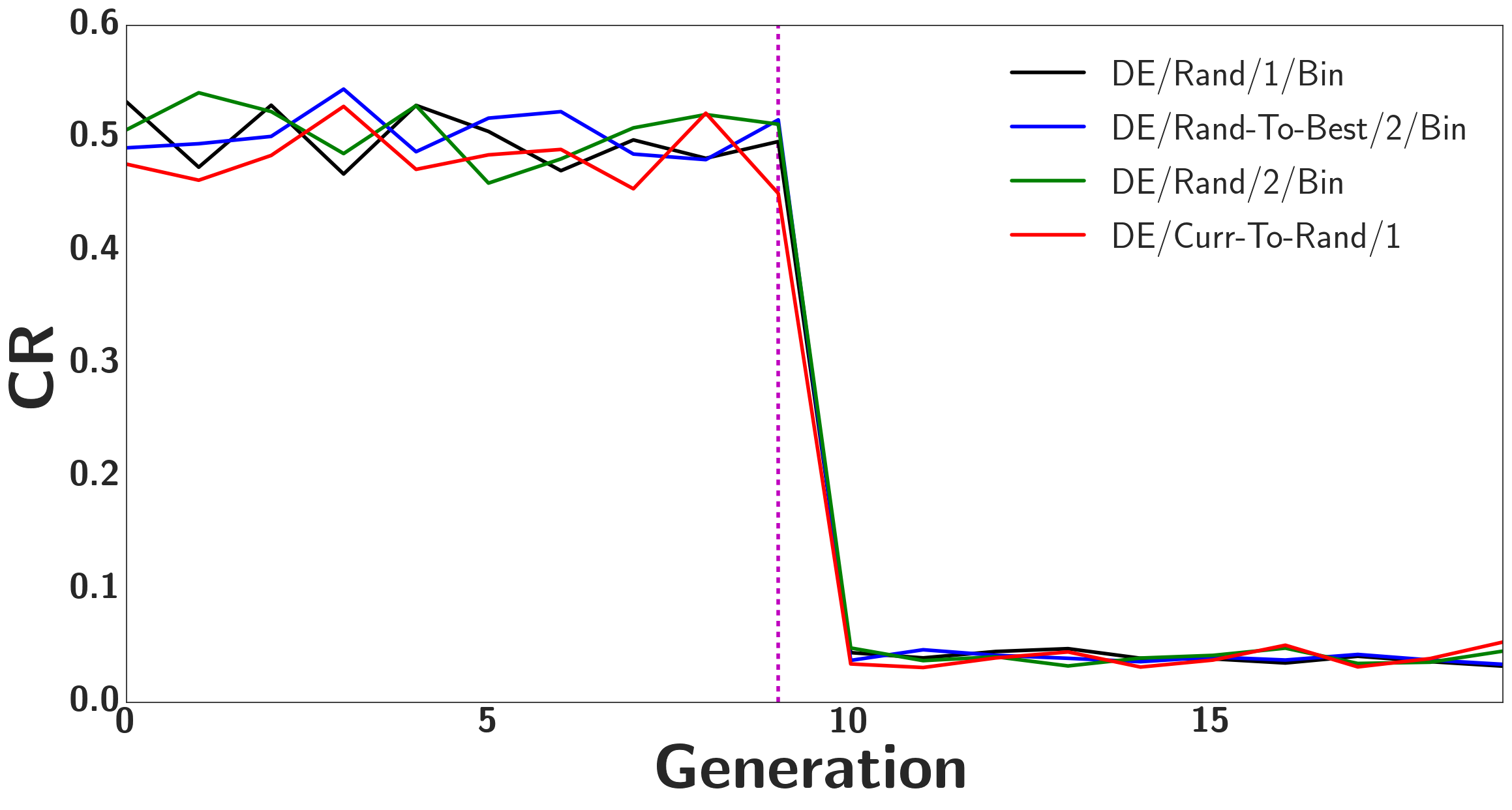}
\caption{}\label{fig:crAVG}
\end{subfigure}
\begin{subfigure}{1\columnwidth}
\centering
\includegraphics[width=1\columnwidth]{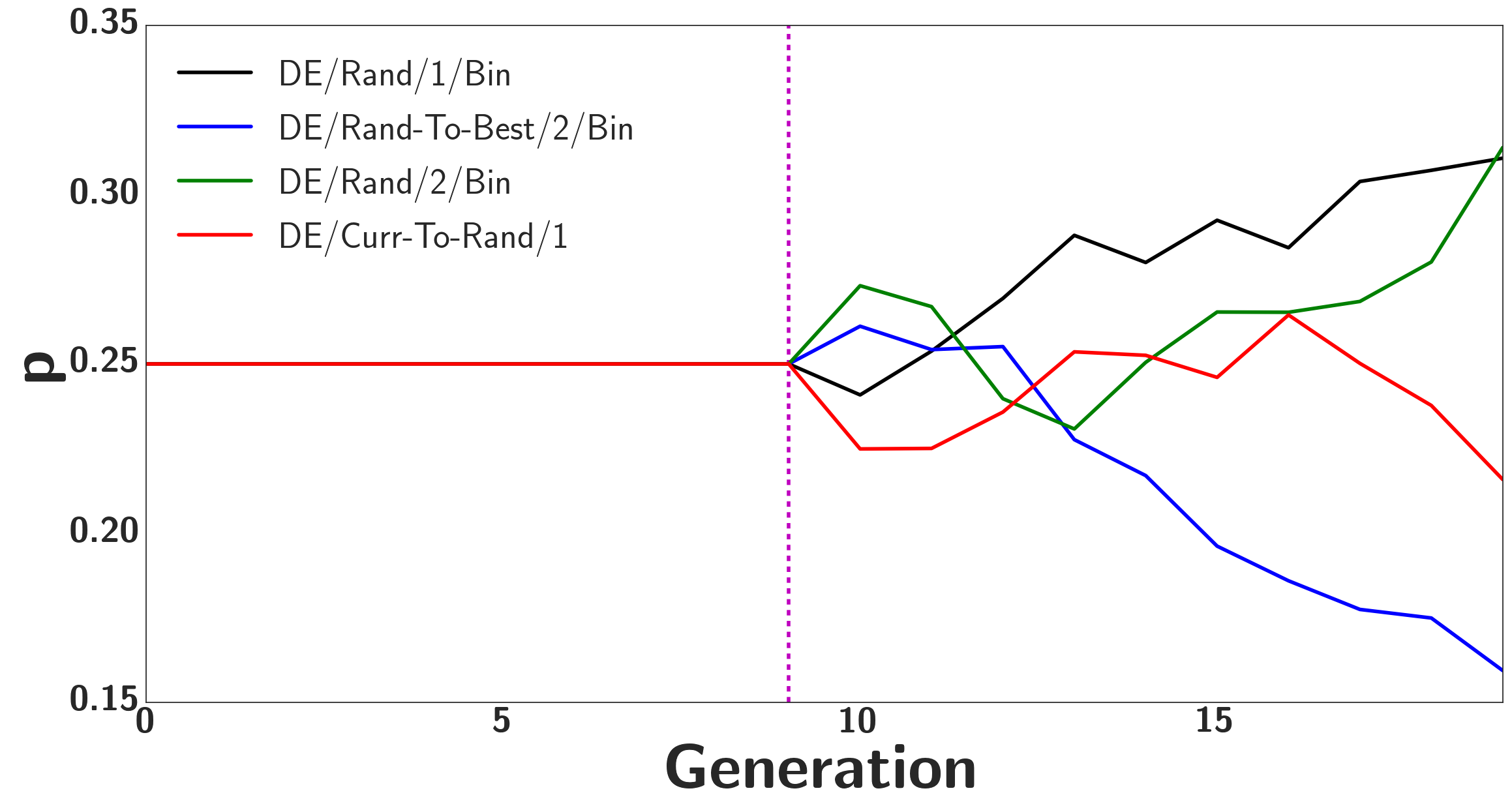}
\caption{}\label{fig:pAVG}
\end{subfigure}
\caption{Averages $F_{acc}(\lambda)$, $F$, $CR$, and $p$ for the SADE population over 19 generations. The dotted vertical line indicates that the learning period has ended.}
\end{figure}

The SADE algorithm performed the best out of all of the algorithms. \fig{fig:sAVG} shows the average $Fitness_{mod}$ of the population over 19 generations. The average $Fitness_{mod}$ converged by five iterations. The max $Fitness_{mod}$ starts off at 0. This indicates that a 100\% accuracy candidate was found in the initialisation period. The max $Fitness_{mod}$ then rises to 400 which is not visible as the range of the average score is -50000 to -1500.\par

The $F$ average results in \fig{fig:fAVG} are quite interesting. They start off at 1 as they are selected from $U([0,2])$ and then drop down to 0.5 as they are selected from $U([0,1])$ after the first generation.  Once the learning period has finished all of the $F$ values have converged to less than 0.1. This indicates that the $F$ values that are having the most success are small and therefore taking advantage of exploration rather than exploitation. It was unexpected that the algorithm would find a min/max within so few generations. This could be why the authors select $F$ from $N(0.5,0.3)$ forcing $F$ to range from -0.4 to 1.4. With $F$ this small the algorithm would effectively be performing gradient descent. However, this could be because the function on the restricted space doesn't have many local maxima. Indeed, these results do come from the best performing LGMD model found.

\fig{fig:crAVG} shows how the CR for each function changes over time.  For the first nine generations, the CR values are selected from $U([0,1])$ and so the mean stays at 0.5. However, as with the $F$ mean values once the learning period is over, all of the CR values go down to less than 0.1. This means that less than 10\% of the mutations will generally take place. From a set of 11 hyper-parameters this means that probabilistically one value will change in addition to the random index that is chosen. $CR$ is generally associated with convergence.

\begin{table}[h!tbp]
\centering
\caption{Parameters used by each model. \label{tab:PARAMS}}
\begin{tabular}{|l||c|c|c|c||}
\hline
 \textbf{Parameter} & \textbf{LGMD} &\textbf{A}&\textbf{P}&\textbf{AP}\\
\hline\hline
 $\mathbf{\tau_e (ms)}$  & 5.87&5.87&5.87&5.87 \\\hline
 $\mathbf{\tau_{iA} (ms)}$  & 3.57&3.57&3.57&3.57 \\\hline
 $\mathbf{\tau_{iB} (ms)}$ & 4.20&4.20&4.20&4.20 \\\hline
 $\mathbf{q_{eP} (pA)}$ & 1014.00&1014.00&1014.00&1014.00 \\\hline
 $\mathbf{q_{eS} (pA)}$ &4635.30&4635.30&4635.30&4635.30  \\\hline
 $\mathbf{q_{eIP} (pA)}$&84.26&84.26 & 84.26& 84.26 \\\hline
 $\mathbf{q_{eIS} (pA)}$ & 168.11&168.11&168.11&168.11 \\\hline
 $\mathbf{q_{eL} (pA)}$ & 80.00& 100.00 &80.00 &100.00  \\\hline
 $\mathbf{inhA_S (1)}$ &1.19&1.19&1.19&1.19\\\hline
 $\mathbf{inhB_S (1)}$ &1.50&1.50&1.50&1.50\\\hline
 $\mathbf{inhA_L (1)}$ 6&0.14&0.14&0.14&0.14 \\\hline
 $\mathbf{a (1)}$ &-&0.79&-&0.79\\\hline
 $\mathbf{b (1)}$ &-&14.51&-&14.51\\\hline
 $\mathbf{\tau_{w_{adapt}} (ms)}$ &-&30.00&-&30.00 \\\hline
 $\mathbf{\tau_{pre} (ms)}$ &-&-&1.56&1.56 \\\hline
 $\mathbf{\tau_{post (ms)}}$ &-&-&10.03&10.03\\\hline
 $\mathbf{\Delta_{pre} (1)}$ &-&-&0.031&0.031\\\hline
 $\mathbf{\Delta_{post} (1)}$ &-&-&0.027&0.027\\\hline
 $\mathbf{c (1)}$ &-&-&0.05&0.05\\\hline
\hline
\end{tabular}
\end{table} 

The probability of each function being chosen is shown in \fig{fig:pAVG}. The probabilities are fixed at 0.25 for the first 9 generations and then they vary based on their success. It is interesting to see that in spite of the $F$ and $CR$ values suggesting that the algorithm is converging on a solution, the DE/Rand-to-Best/2/Bin algorithm is the least successful. The DE/Curr-to-Rand/1 algorithm performs relatively well until about 16 generations where it tapers off. The DE/Rand/2/bin algorithm dips initially but then increases as DE/Curr-to-Rand/1 starts to drop off. The DE/Rand/1/bin remains relatively high during the entire algorithm only to be overtaken by The DE/Rand/2/bin in the last generation.

\subsection{Comparison of Models}
\label{ssec:modcomp}
\tab{tab:PARAMS} shows the selected final parameters of each model. These values were all found by the SADE algorithm, due to the superior quality of its results. The (1) tag in the parameter column indicates that the variable is unit-less.

In both models with plasticity, the clamping value $c$ was set to 0.05, or 5\%.

\begin{table*}[h!tbp] 
\centering
\caption{Quality metrics of the performance of different LGMD models for different simulated looming stimuli. }\label{tab:LGMDComp}
\begin{tabular}{|>{\raggedright\arraybackslash}p{85pt}|>{\centering\arraybackslash}p{50pt}||>{\centering\arraybackslash}p{52pt}|>{\centering\arraybackslash}p{58pt}|>{\centering\arraybackslash}p{52pt}|>{\centering\arraybackslash}p{58pt}||}\hline
\textbf{Stimulus}&\textbf{Model}&\textbf{Accuracy}&\textbf{Sensitivity}&\textbf{Precision}&\textbf{Specificity}\\\hline\hline
\multirow{5}{*}{\textbf{composite}}
&\textbf{LGMD}&0.90&1.00&0.83&0.80\\\cline{2-6}
&\textbf{A}&0.90&1.00&0.83&0.80\\\cline{2-6}
&\textbf{P}&0.90&1.00&0.83&0.80\\\cline{2-6}
&\textbf{AP}&0.90&1.00&0.83&0.80\\\hline
\multirow{5}{*}{\textbf{circleSlow}}
&\textbf{LGMD}&0.80&0.60&1.00&1.00\\\cline{2-6}
&\textbf{A}&0.80&0.60&1.00&1.00\\\cline{2-6}
&\textbf{P}&0.90&0.80&1.00&1.00\\\cline{2-6}
&\textbf{AP}&1.00&1.00&1.00&1.00\\\hline
\multirow{5}{*}{\textbf{circleFast}}
&\textbf{LGMD}&1.00&1.00&1.00&1.00\\\cline{2-6}
&\textbf{A}&1.00&1.00&1.00&1.00\\\cline{2-6}
&\textbf{P}&1.00&1.00&1.00&1.00\\\cline{2-6}
&\textbf{AP}&1.00&1.00&1.00&1.00\\\hline
\multirow{5}{*}{\textbf{squareSlow}}
&\textbf{LGMD}&1.00&1.00&1.00&1.00\\\cline{2-6}
&\textbf{A}&1.00&1.00&1.00&1.00\\\cline{2-6}
&\textbf{P}&1.00&1.00&1.00&1.00\\\cline{2-6}
&\textbf{AP}&1.00&1.00&1.00&1.00\\\hline
\multirow{5}{*}{\textbf{squareFast}}
&\textbf{LGMD}&1.00&1.00&1.00&1.00\\\cline{2-6}
&\textbf{A}&1.00&1.00&1.00&1.00\\\cline{2-6}
&\textbf{P}&1.00&1.00&1.00&1.00\\\cline{2-6}
&\textbf{AP}&1.00&1.00&1.00&1.00\\\hline
\end{tabular}
\end{table*}

\begin{figure}[t!p]
\centering
\begin{subfigure}{1\columnwidth}
\centering
\includegraphics[width=1\columnwidth]{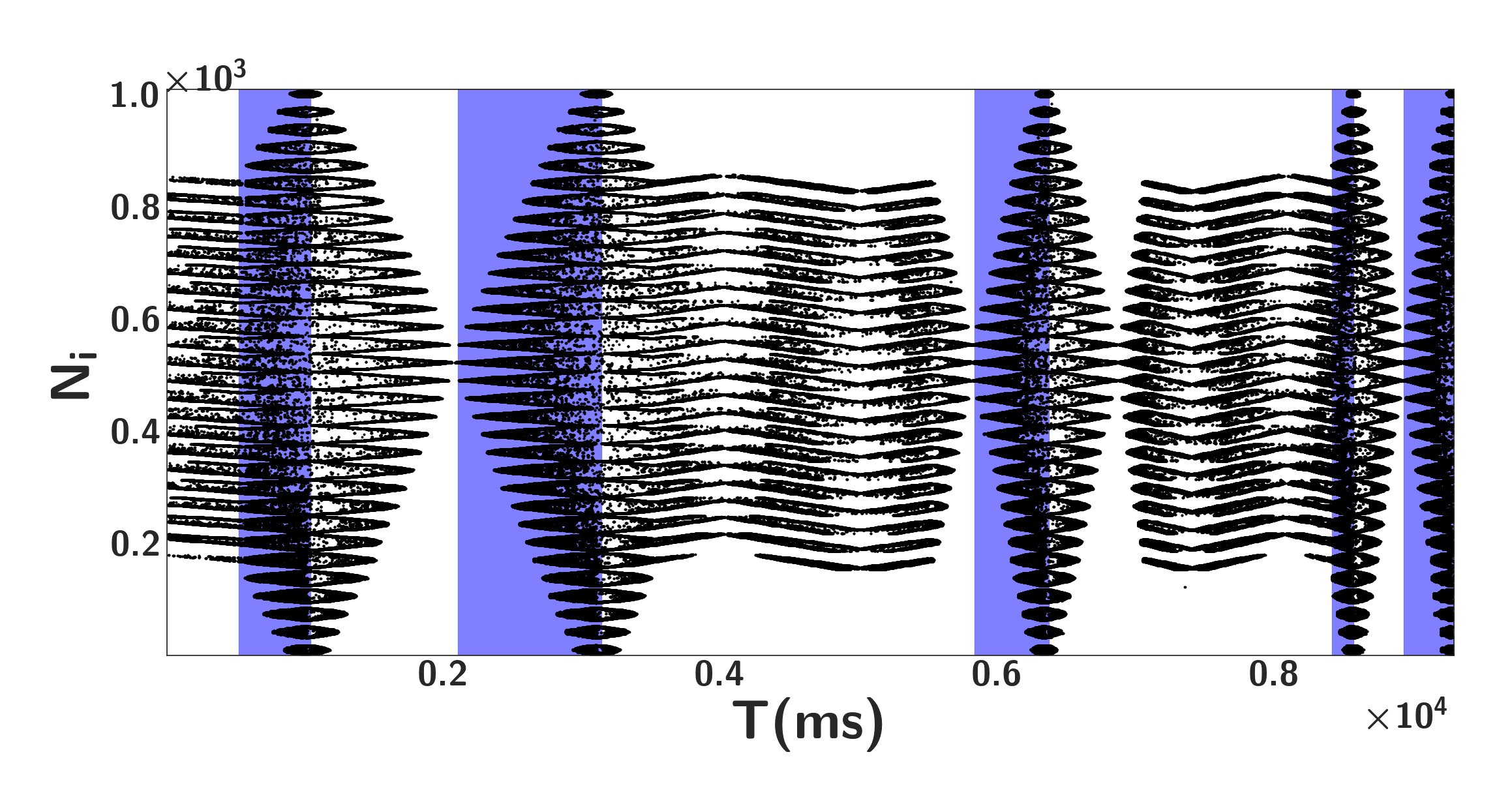}
\caption{Filtered Composite Input (P Layer Raster Plot).}\label{fig:comP}
\end{subfigure}
\begin{subfigure}{1\columnwidth}
\centering
\includegraphics[width=1\columnwidth]{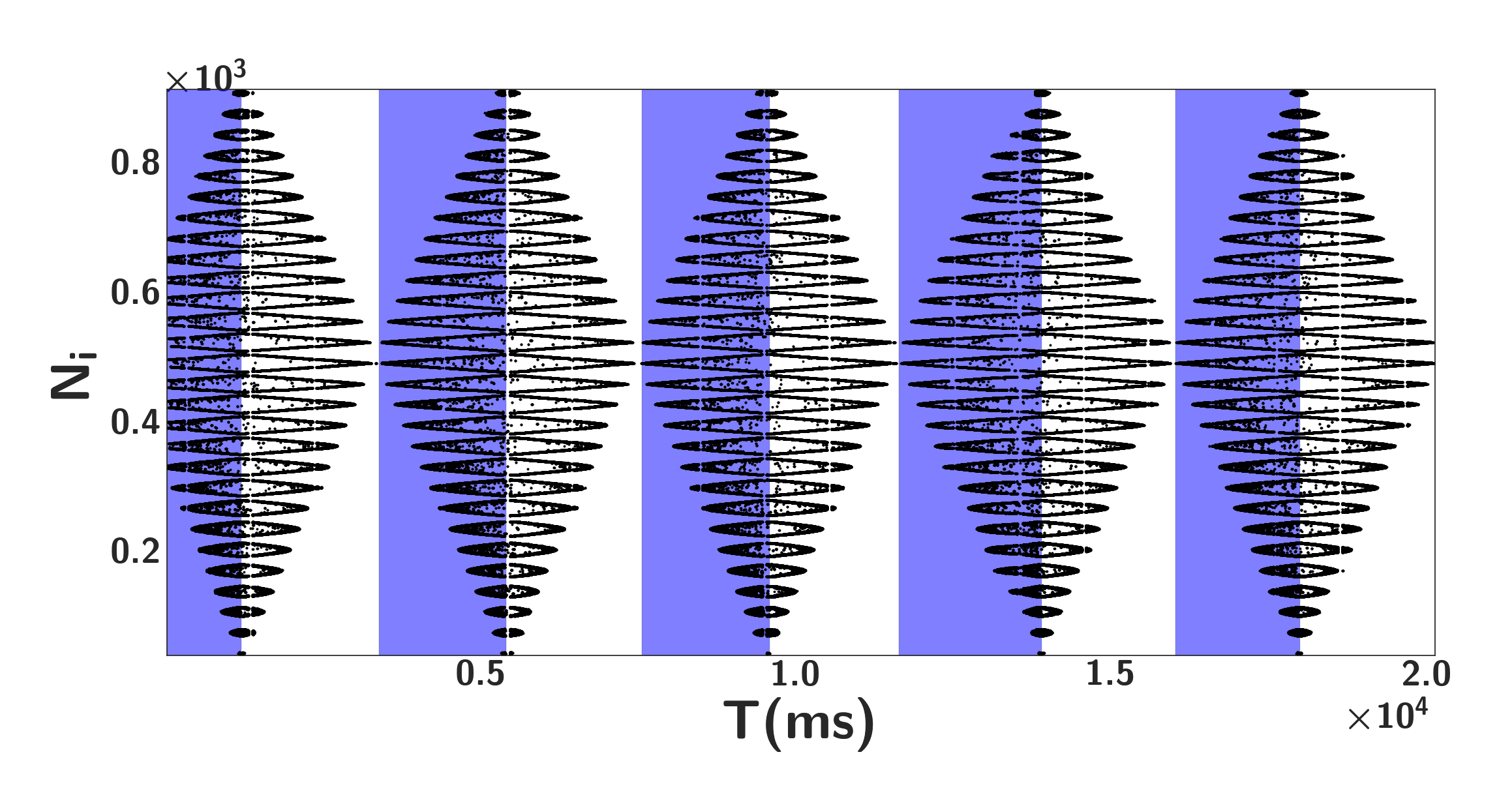}
\caption{Filtered circleSlow Input (P Layer Raster Plot).}\label{fig:cSP}
\end{subfigure}
\begin{subfigure}{1\columnwidth}
\centering
\includegraphics[width=1\columnwidth]{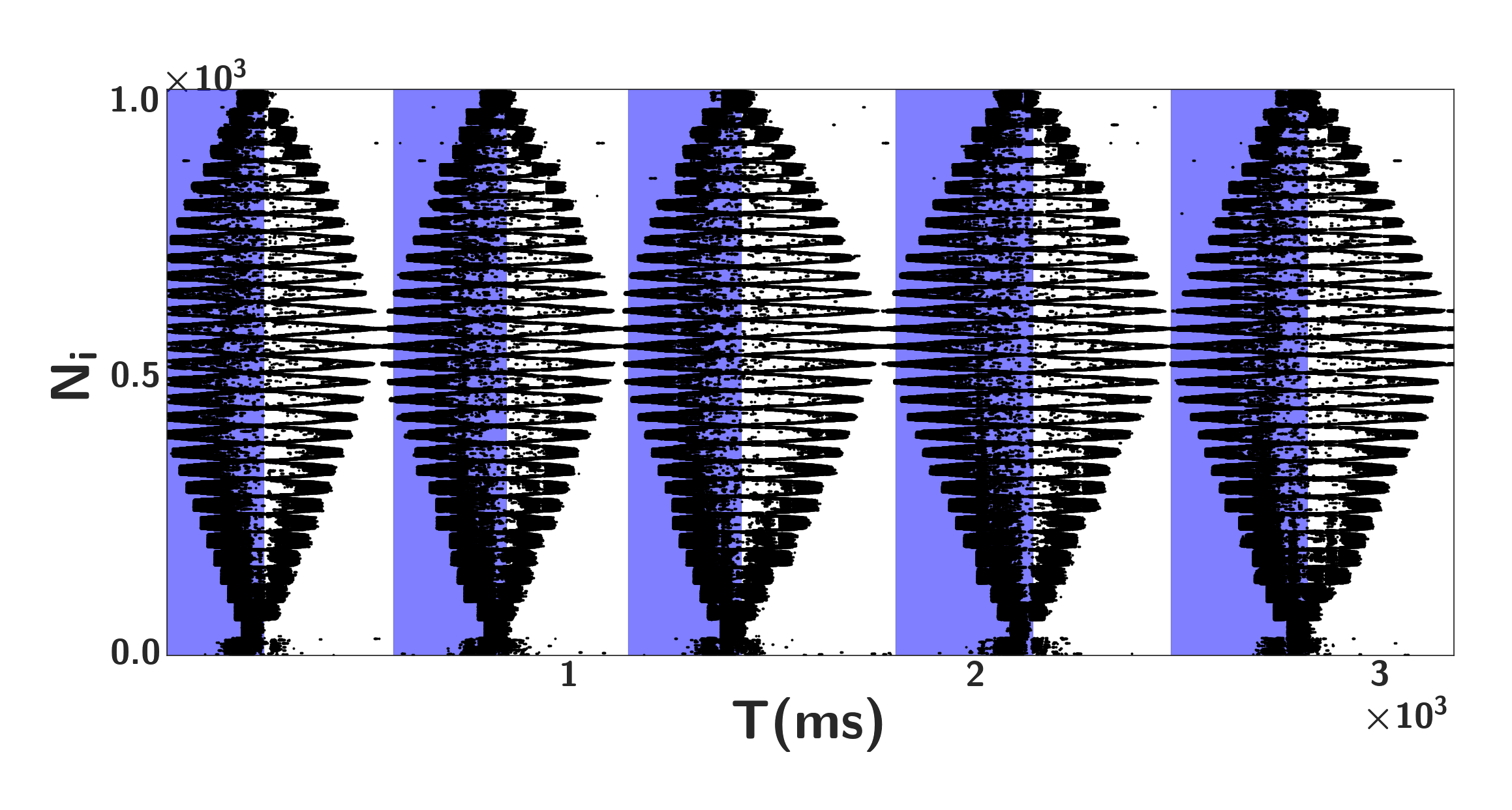}
\caption{Filtered squareFast Input (P Layer Raster Plot).}\label{fig:sFP}
\end{subfigure}
\caption{The input layer for the simple stimuli.The white and coloured backgrounds indicate non-looming and looming respectively.}\label{simpleStim}
\end{figure}

As expected, all of the models have a $\tau_{iA} < \tau_{iB}$ which means that the $B$ inhibitions will persist for longer and have slower dynamics relative to the $A$ inhibitions. What is unexpected is that the $B$ inhibitions also have stronger current injection than the $A$ inhibitions. On top of this, both of the inhibitory current injections are actually stronger than the excitatory connections. Whereas the model in~\cite{yue2010reactive} with discrete dynamics had relatively low inhibitory current injections, with $inhA_S = 0.25$ and $inhB_S = 0.125$ of the excitation strength. Clearly, there is a difference between the neuron models that are used, but this is an interesting outcome nonetheless.

\begin{figure}[h!tpb]
\centering
\begin{subfigure}{1\columnwidth}
\centering
\includegraphics[width=1\columnwidth]{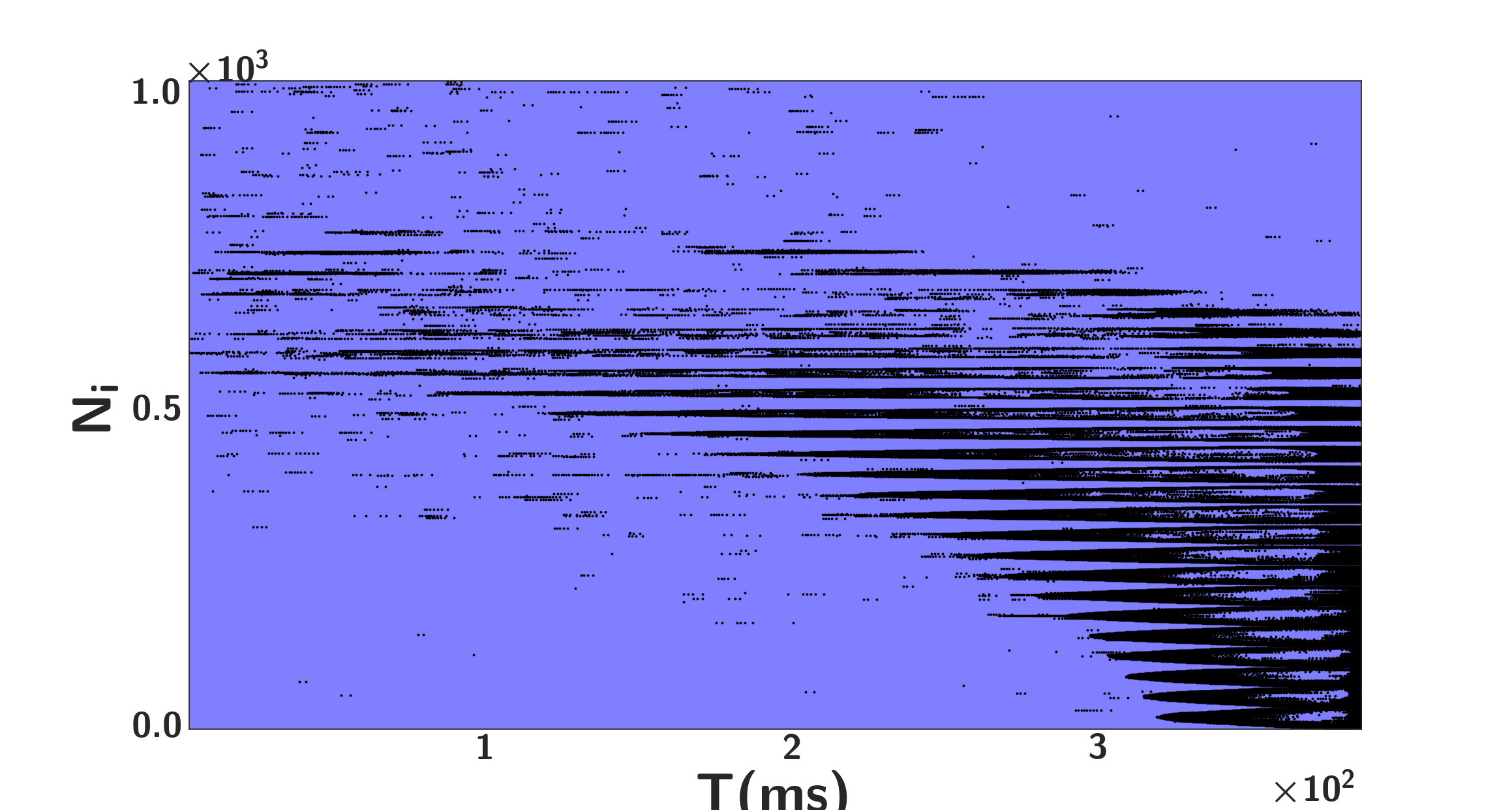}
\caption{Filtered ballRoll2 Input (P Layer Raster Plot).}\label{fig:BRP}
\end{subfigure}
\begin{subfigure}{1\columnwidth}
\centering
\includegraphics[width=1\columnwidth]{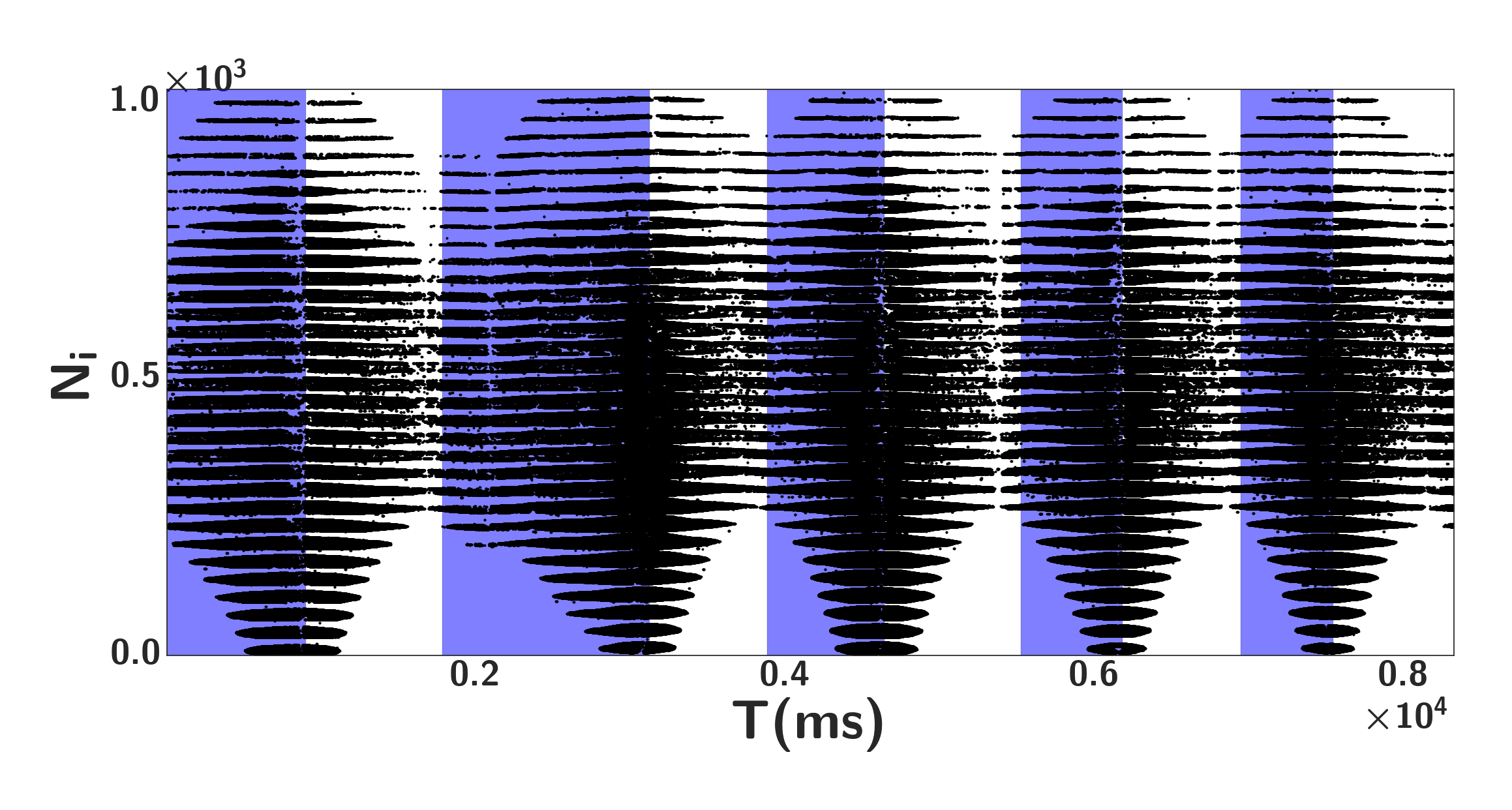}
\caption{Filtered cupQUAV Input (P Layer Raster Plot).}\label{fig:lCP}
\end{subfigure}
\begin{subfigure}{1\columnwidth}
\centering
\includegraphics[width=1\columnwidth]{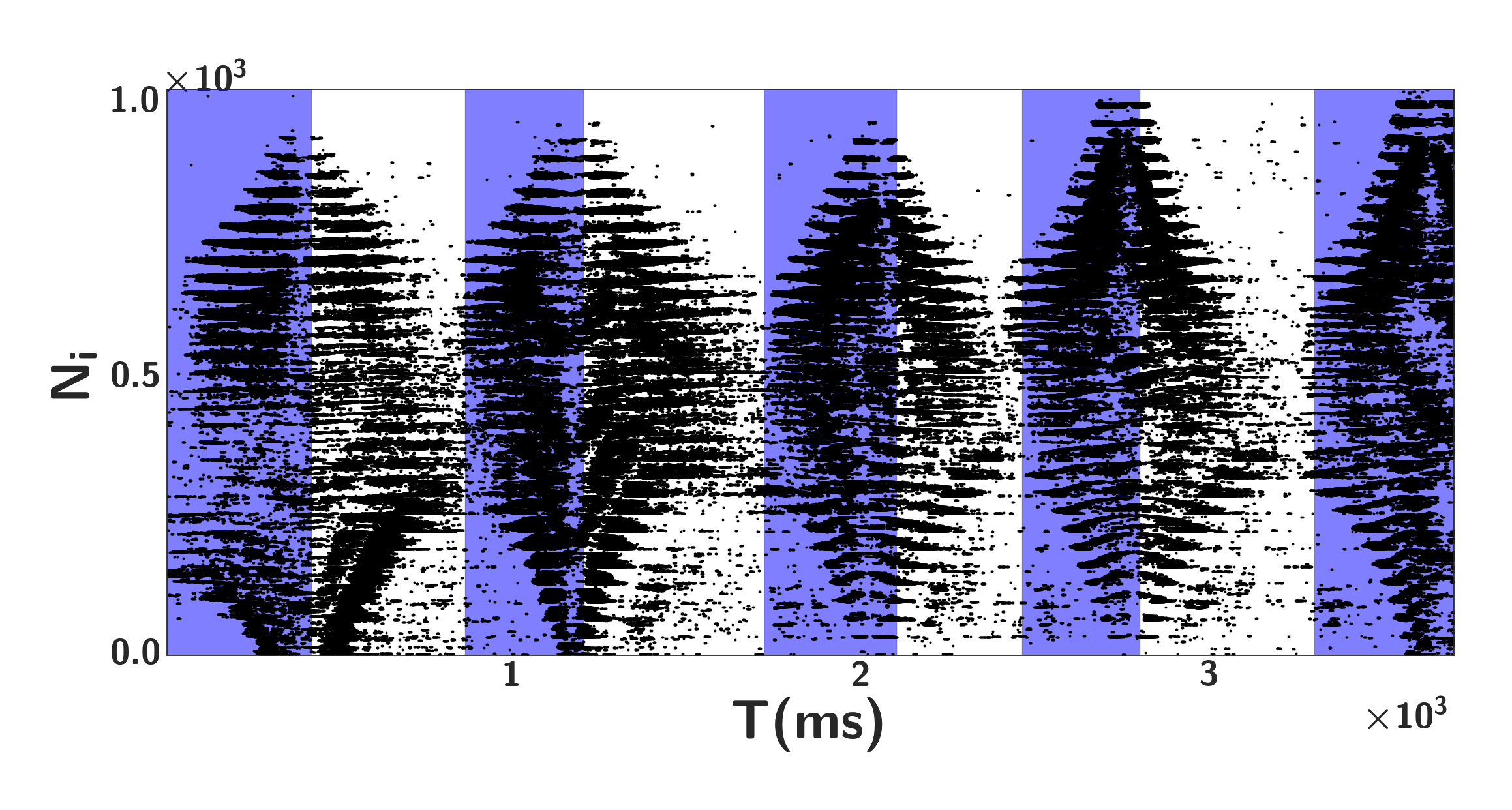}
\caption{Filtered Hand Input (P Layer Raster Plot).}\label{fig:hP}
\end{subfigure}
\caption{Complex real stimuli. The white and coloured backgrounds indicate non-looming and looming respectively.}
\end{figure}

\tab{tab:LGMDComp} shows the accuracy, sensitivity, precision, and specificity for each LGMD model for a given simple stimulus. The stimuli can be described as follows: 

\begin{description}[topsep=0.2ex]
\item[\textbf{composite:}] A standard test bench stimulus that consists of a black circle on a white background that translates and looms at increasing speeds. \fig{fig:comP} shows the composite input. 
\item[\textbf{circleFast/Slow:}] A purely looming black circle on white background at high or low speeds. Collected on hovering QUAV. \fig{fig:cSP} shows the circleFast/Slow stimulus. 
\par
\item[\textbf{squareFast/Slow:}] A purely looming black square on a white background at high/low speeds. \fig{fig:sFP} shows the squareFast/Slow stimulus.
\end{description}

The results in \tab{tab:LGMDComp} show that the models performed well ($Accuracy\geq 0.8$) on most of the stimuli. LGMD and \textbf{A} perform poorly on the circleSlow test, missing two out of five of the looming stimuli. \textbf{P} misses one looming stimulus, and \textbf{AP} detects all stimuli accurately. The plasticity increases the weights of important connections and the adaptation filters out over excited neurons.

These results show that the models are capable of detecting looming stimuli of varying speeds and of differentiating between translation and looming stimuli for the most part. \textbf{AP} scored 100\% in every test besides the composite stimulus where it misclassified the first short translation as a loom. This can probably be attributed to the network not starting in its resting/equilibrium state. 

After performing the simulated experiments of computer generated shapes, real objects moving towards and away from the camera were recorded. These stimuli can be described as:
\begin{description}
\item[\textbf{ballRoll[1-3]:}] Three different runs of a white ball rolling towards the camera on a black platform at different angles and speeds. This is a purely looming stimulus. \fig{fig:BRP} shows one of the three ball rolls.
\item[\textbf{cupQUAV:} ] A QUAV flying towards a cup suspended in front of it with a white wall behind it. This is a self stimulus.\fig{fig:lCP} shows the QUAV cup stimulus.
\item[\textbf{Hand:} ] A Hand moving towards and away from the hovering QUAV.  \fig{fig:hP} shows the looming hand stimulus. 
\end{description}\par

\fig{fig:BRP}, \fig{fig:lCP}, and \fig{fig:hP} show that the real stimuli tend to have more noise and do not adhere to a strong pattern when compared to \fig{fig:comP}, \fig{fig:cSP}, and \fig{fig:sFP}. \tab{tab:LGMDComp2} shows that the models do not perform as well on real world stimuli. ballRoll[1-3] is the simplest real stimulus, and as such \textbf{P} and \textbf{AP} achieved full accuracy. LGMD and \textbf{A} missed one roll.

\begin{table*}[h!tbp]
\centering
\caption{Quality metrics of the performance of different LGMD models for different real looming stimuli}\label{tab:LGMDComp2}
\begin{tabular}{|>{\raggedright\arraybackslash}p{85pt}|>{\centering\arraybackslash}p{50pt}||>{\centering\arraybackslash}p{52pt}|>{\centering\arraybackslash}p{58pt}|>{\centering\arraybackslash}p{52pt}|>{\centering\arraybackslash}p{58pt}||}\hline
\textbf{Stimulus}&\textbf{Model}&\textbf{Accuracy}&\textbf{Sensitivity}&\textbf{Precision}&\textbf{Specificity}\\\hline\hline
\multirow{5}{*}{\textbf{ballRoll[1-3]}}
&\textbf{LGMD}&0.66&0.66&1.00&0.00\\\cline{2-6}
&\textbf{A}&0.66&0.66&1.00&0.00\\\cline{2-6}
&\textbf{P}&1.00&1.00&1.00&0.00\\\cline{2-6}
&\textbf{AP}&1.00&1.00&1.00&0.00\\\hline
\multirow{5}{*}{\textbf{cupQUAV}}
&\textbf{LGMD}&0.70&1.00&0.62&0.40\\\cline{2-6}
&\textbf{A}&0.70&1.00&0.62&0.40\\\cline{2-6}
&\textbf{P}&0.70&1.00&0.62&0.40\\\cline{2-6}
&\textbf{AP}&0.80&1.00&0.71&0.60\\\hline
\multirow{5}{*}{\textbf{hand}}
&\textbf{LGMD}&0.50&1.00&0.50&0.00\\\cline{2-6}
&\textbf{A}&0.50&1.00&0.50&0.00\\\cline{2-6}
&\textbf{P}&0.50&1.00&0.50&0.00\\\cline{2-6}
&\textbf{AP}&0.50&1.00&0.50&0.00\\\hline
\end{tabular}
\end{table*}

Surprisingly good results come from the cupQUAV stimulus: 70\% accuracy for all models except for \textbf{AP}, which had 80\%. It is worth noting that \textbf{AP} performed consistently well when compared with the other models.

The possibility of detecting the hand by stochastically dropping pixel-events, was investigated. Dropping 50\% of the DVS events and re-optimising the network gave 100\% accuracy for the hand and cupQuad stimulus. However, in doing this, the network was no longer robust to the speed changes in the composite benchmark test. Indeed, even using all of the pixels, the network could be optimised to work on the real world stimuli. The inhibition values went up and the gain values went down, meaning the network struggled to spike on stimuli that weren't noisy or event heavy. Some sort of additional pre-filtering could be useful in getting the looming network to be fully robust in all situations. 

\subsubsection{The Effect of Changing $c$ on Plasticity}
\label{ssec:plasC}

\begin{figure}[h!tbp]
\begin{subfigure}{1\columnwidth}
\centering
\includegraphics[width=1\columnwidth]{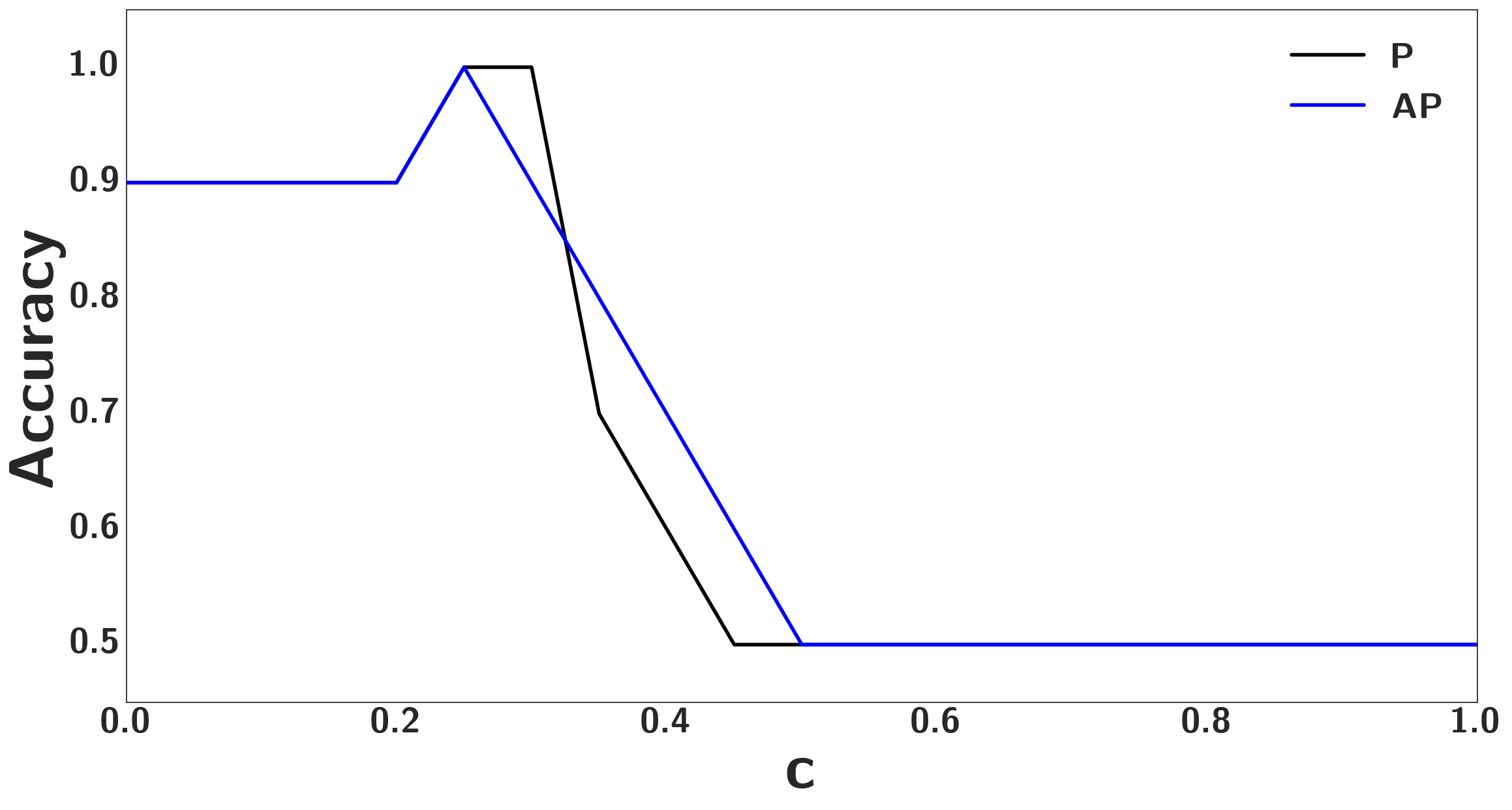}
\caption{Effect of changing the $c$ clamping value on the learning weight $w$ for the composite stimulus}\label{fig:pCom}
\end{subfigure} 
\begin{subfigure}{1\columnwidth}
\centering
\includegraphics[width=1\columnwidth]{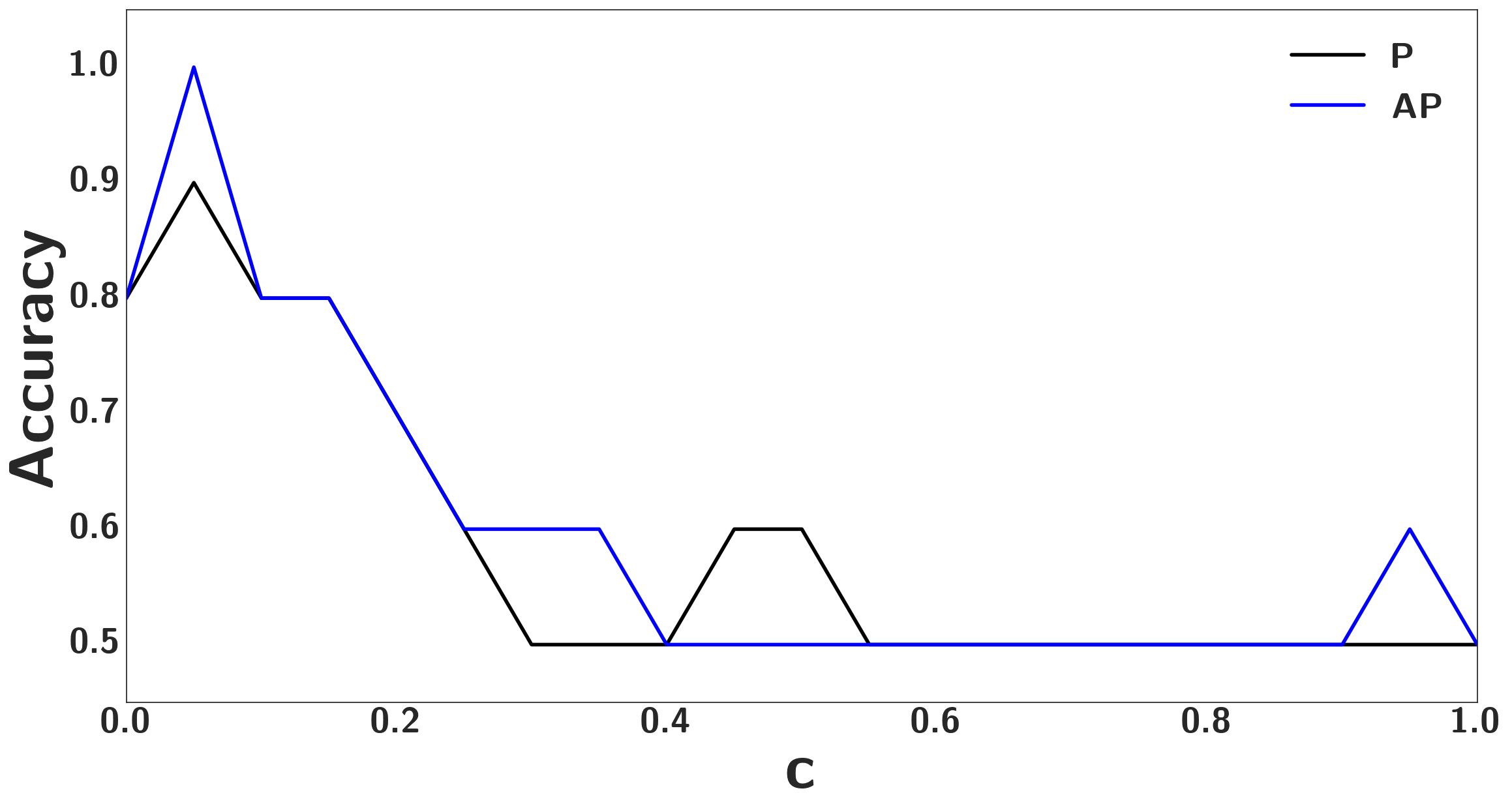}
\caption{Effect of changing the $c$ clamping value on the learning weight $w$ for the circleSlow stimulus}\label{fig:psloC}
\end{subfigure} 
\begin{subfigure}{1\columnwidth}
\centering
\includegraphics[width=1\columnwidth]{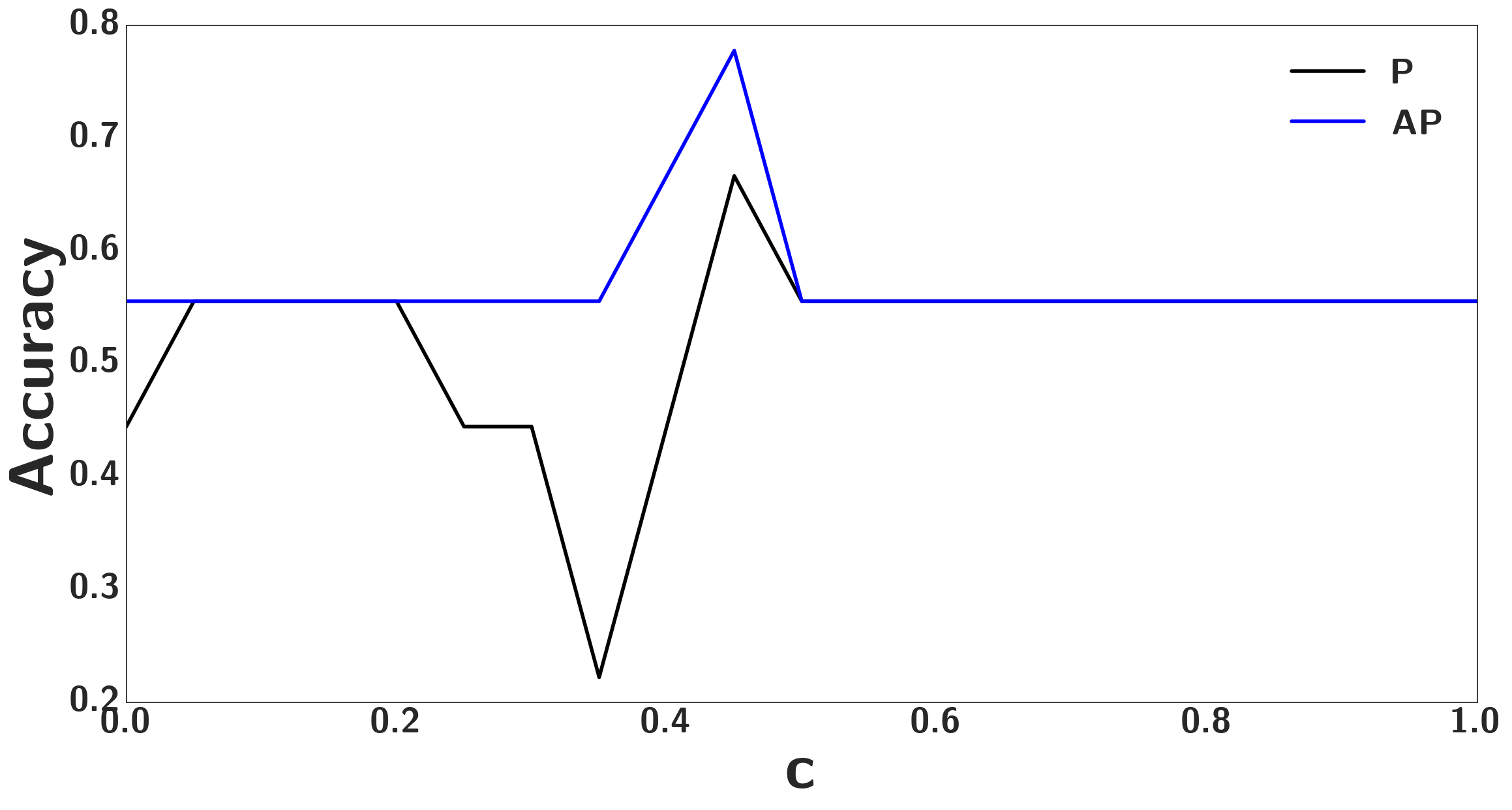}
\caption{Effect of changing the $c$ clamping value on the learning weight $w$ for the hand stimulus}\label{fig:pH}
\end{subfigure}
\caption{The effect of changing the clamping value on various stimuli}
\end{figure}

\fig{fig:pCom}, \fig{fig:psloC}, and \fig{fig:pH} show how changing the bounds of the plasticity clamping changes the LGMD (\textbf{P} model) accuracy for the composite, cicleSlow, and hand stimuli respectively.

Interestingly, for the two simulated stimuli increasing the clamping to beyond 25\% caused the accuracy to drop to 50\%. The sensitivity dropped to 0\% indicating that it was no longer detecting looms and that the synaptic weights were no longer causing the LGMD neuron to fire.

Increasing the clamping to 45\% increases the accuracy for both the \textbf{P} and \textbf{AP} models on the hand stimulus. This shows that plasticity is a double edged sword that can both improve and degrade the performance of the model. Knowledge about the nature of your input can help to determine what level of plasticity you require. In all cases, a small contribution of plasticity improved the performance. This could be due to the fact that the amount of noise in the simulated stimuli was far less than the noise in the real stimuli.

\subsubsection{Weight Visualisation}

\begin{figure}[h!tbp]
\centering
\includegraphics[width=1\columnwidth]{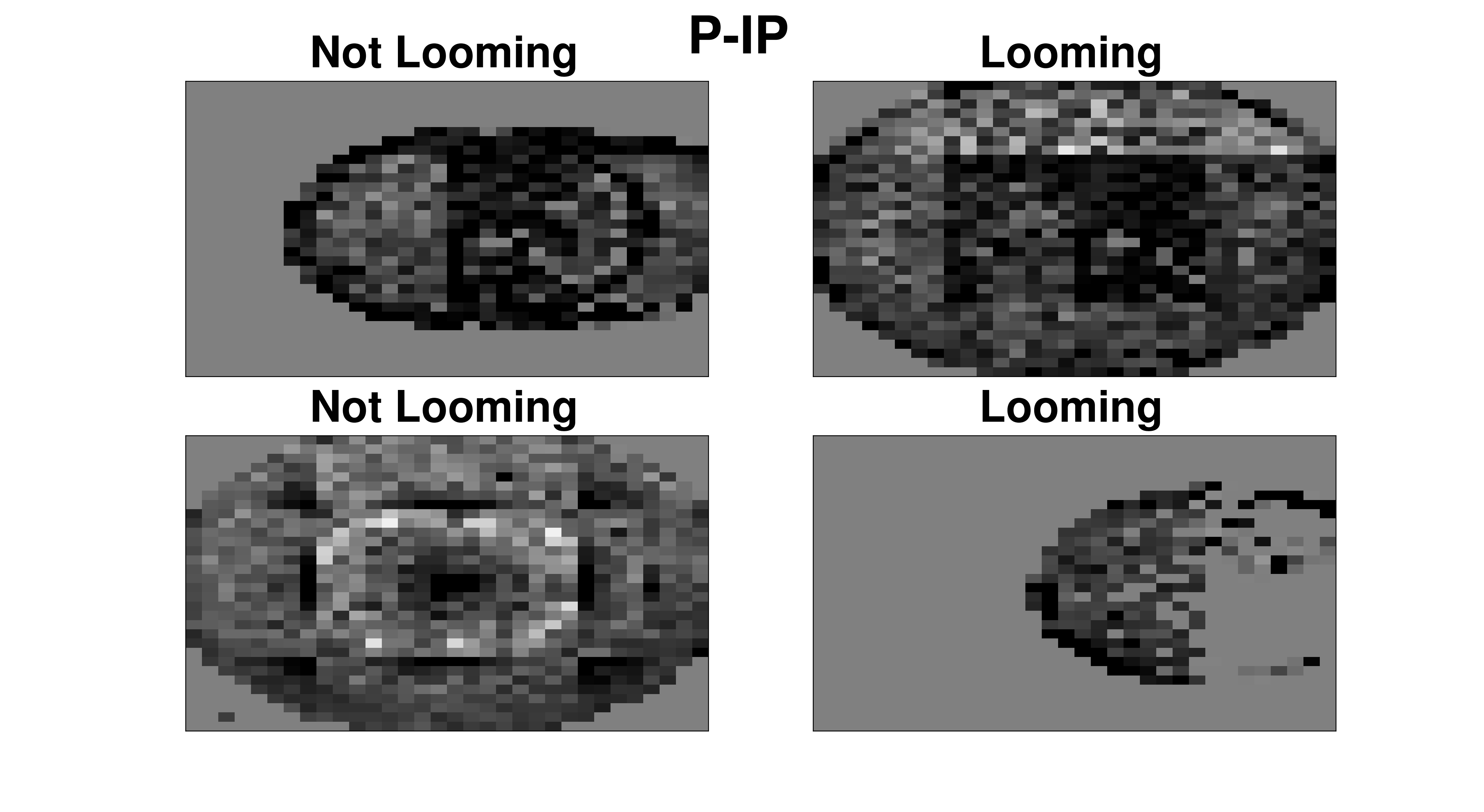}
\caption{The weights of the synapses from the P to the IP layer at the end of a looming or non-looming sequence.}\label{fig:PIP}
\end{figure}

\begin{figure}[h!tbp]
\centering
\includegraphics[width=1\columnwidth]{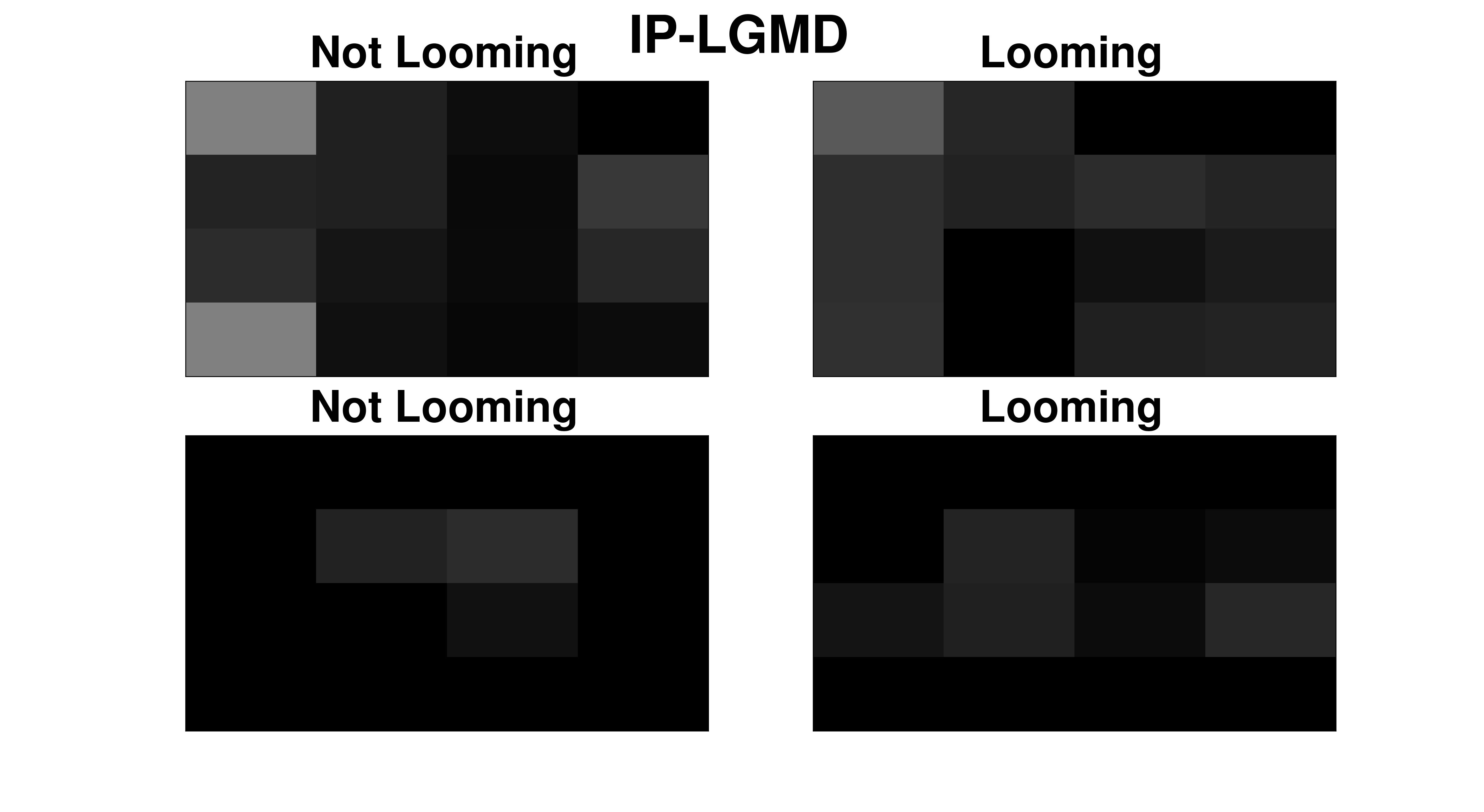}
\caption{The weights of the synapses from the IP to the LGMD layer at the end of a looming or non-looming sequence.}\label{fig:IPL}
\end{figure}

\begin{figure}[h!tbp]
\centering
\includegraphics[width=1\columnwidth]{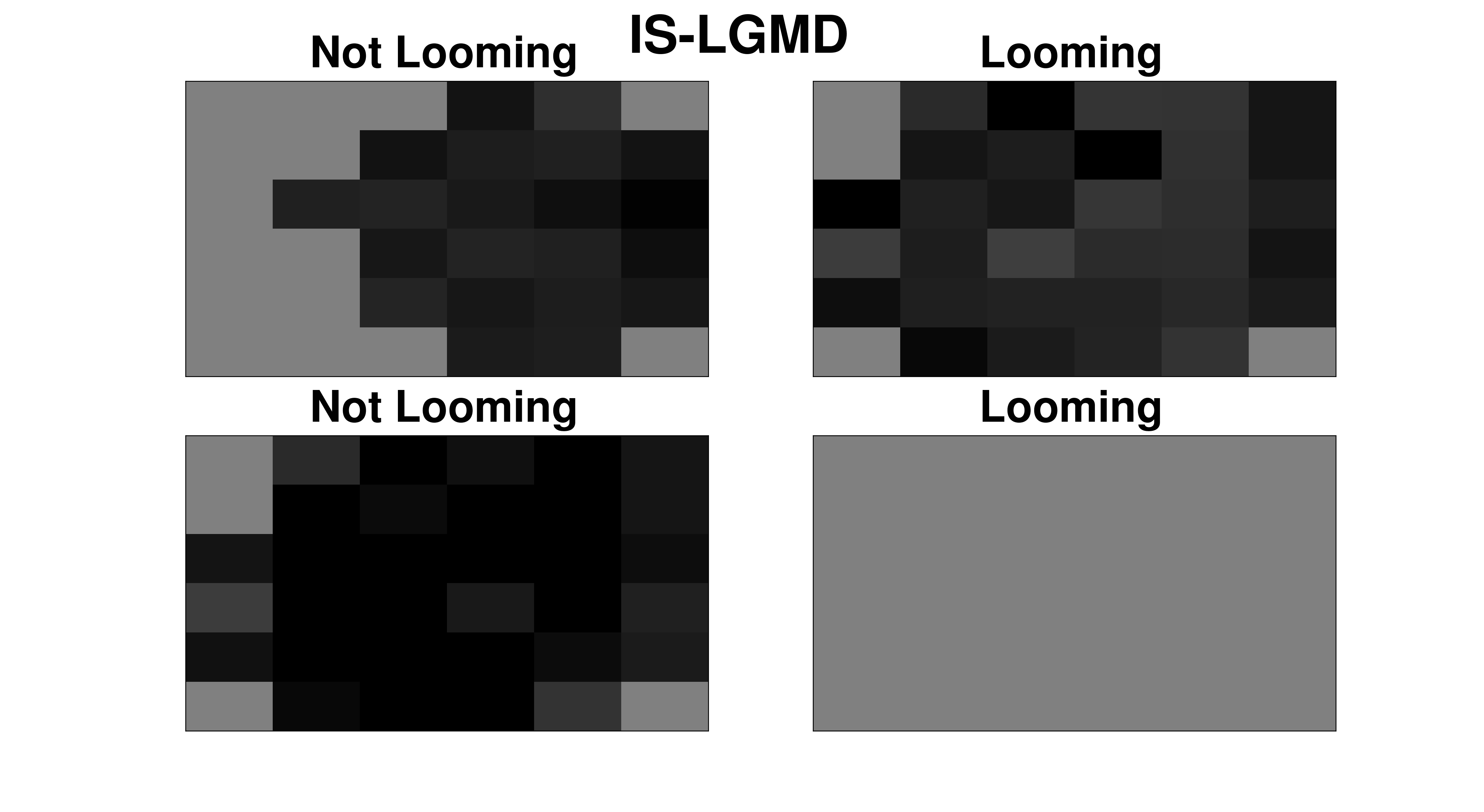}
\caption{The weights of the synapses from the IS to the LGMD layer at the end of a looming or non-looming sequence.}\label{fig:ISL}
\end{figure}

\fig{fig:PIP}, \fig{fig:IPL}, and \fig{fig:ISL} show snapshots of the weights at the end of each looming or non-looming sequence. We used the \textbf{P} model with $c=0.25$ on the composite stimulus. This was done because it achieved 100\% accuracy and 25\% clamping has greater weight variation than 10\%. 

\fig{fig:PIP} is interesting as it most obviously correlates to the input. We can see in the first non-looming snapshot that the P-IP layer is strongly inhibiting a circle translating from right to left. In the looming section, the circle is moving outwards and the central weights have the highest density of low values. This shows that the centre of the circle is not associated with the output. In the second non-looming snapshot, the density of high values is in the centre of the circle showing that it has higher inhibitions. 

\fig{fig:IPL} and \fig{fig:ISL} show the IP and IS connections to the LGMD layer. The IP-LGMD snapshots tend to have higher weights during looming than non-looming stimuli. Interestingly, in both figures, the highest value is one, meaning that the weights have only become weaker than they initially were, at least for these selected times. 

\section{Conclusions}
We implemented a neuromorphic model of the locust LGMD network using recordings from a UAV equipped with a DVS sensor as inputs. The neuromorphic LGMDNN was capable of differentiating between looming and non-looming stimuli. It was capable of detecting the black and white simple stimuli correctly regardless of speed and shape. Real-world stimuli performed relatively well using the parameters found by the optimiser for synthesised stimuli. However, when re-optimised, the real-world stimuli performed comparably to the synthesised stimuli. This was mainly because real-world stimuli tend to contain a higher number of luminance changes and therefore the magnitude parameters needed to be reduced. 

We showed that BO, DE, and SADE are capable of finding parameter values that give the desired performance in the LGMDNN model. It can be seen that SADE statistically significantly outperformed DE on all metrics besides sensitivity and the number of evaluations,  although the only metrics that formed part of the objective function were fitness and accuracy. Once a suitable objective function was found that accurately described the desired output of the LGMDNN, BO, DE and SADE outperformed hand-crafted attempts. The algorithms were able to achieve 100\% accuracy on  black and white simple stimuli of varying shapes and speeds. SADE performed well in this task and we have shown that it is suitable for the optimisation of a multi-layered LGMD spiking neural network. This could save time when developing biologically plausible SNNs in related applications.

In the future, we would like to apply the optimisation algorithms directly to tuning the neuromorphic processors using the neuromorphic model, with the end goal being a closed loop control system on a UAV. Showing that optimisation is effective for selecting parameters on neuromorphic hardware will increase their usability. 


%

\ifCLASSOPTIONcompsoc
  \section*{Acknowledgments}
\else
  \section*{Acknowledgment}
\fi

We are grateful to Prof. Claire Rind, who provided valuable comments and feedback on the definition of the neuromorphic model, and acknowledge the CapoCaccia Cognitive Neuromorphic Engineering workshop, where these discussions and model developments took place.

We would also like to thank INILABs for use of the DVS sensor and the Institute of Neuroinfoamrtics (INI), University of Zurich and ETH Zurich for its neuromorphic processor developments.
Part of this work was funded by the EU ERC Grant ``neuroP'' (257219) and EU H2020-MSCA-IF-2015 grant ``ECogNet'' (707373). 

\ifCLASSOPTIONcaptionsoff
  \newpage
\fi



%
\bibliographystyle{IEEEtran}
\bibliography{TNNLS,biblio}

%




\end{document}